\documentclass[letterpaper]{article} % DO NOT CHANGE THIS
\usepackage{aaai2026}  % DO NOT CHANGE THIS
\usepackage{times}  % DO NOT CHANGE THIS
\usepackage{helvet}  % DO NOT CHANGE THIS
\usepackage{courier}  % DO NOT CHANGE THIS
\usepackage[hyphens]{url}  % DO NOT CHANGE THIS
\usepackage{graphicx} % DO NOT CHANGE THIS
\usepackage{natbib}  % DO NOT CHANGE THIS AND DO NOT ADD ANY OPTIONS TO IT
\usepackage{caption} % DO NOT CHANGE THIS AND DO NOT ADD ANY OPTIONS TO IT
\usepackage{algorithm}
\usepackage{algorithmic}

\usepackage{newfloat}
\usepackage{listings}
\DeclareCaptionStyle{ruled}{labelfont=normalfont,labelsep=colon,strut=off} % DO NOT CHANGE THIS
\floatstyle{ruled}
\newfloat{listing}{tb}{lst}{}
\floatname{listing}{Listing}
\usepackage{multirow} %新增
\usepackage{makecell} %新增
\usepackage{amsmath}
\usepackage{amsfonts}

\title{Spatio-Temporal Context Learning with Temporal Difference Convolution for Moving Infrared Small Target Detection}
\author{
    Houzhang Fang\textsuperscript{\rm 1}\thanks{Corresponding author.},
    Shukai Guo\textsuperscript{\rm 1},
    Qiuhuan Chen\textsuperscript{\rm 1},
    Yi Chang\textsuperscript{\rm 2},
    Luxin Yan\textsuperscript{\rm 2}\\
}
\affiliations{
    \textsuperscript{\rm 1}School of Computer Science and Technology, Xidian University, China\\
    \textsuperscript{\rm 2}School of Artificial Intelligence and Automation, Huazhong University of Science and Technology, China\\
houzhangfang@outlook.com,shukaiguo@stu.xidian.edu.cn,cqh@stu.xidian.edu.cn,\{yichang,yanluxin\}@hust.edu.cn
}

\usepackage{bibentry}
\begin{document}

\maketitle

\begin{abstract}
Moving infrared small target detection (IRSTD) plays a critical role in practical applications, such as surveillance of unmanned aerial vehicles (UAVs) and UAV-based search system. Moving IRSTD still remains highly challenging due to weak target features and complex background interference. Accurate spatio-temporal feature modeling is crucial for moving target detection, typically achieved through either temporal differences or spatio-temporal (3D) convolutions. Temporal difference can explicitly leverage motion cues but exhibits limited capability in extracting spatial features, whereas 3D convolution effectively represents spatio-temporal features yet lacks explicit awareness of motion dynamics along the temporal dimension. In this paper, we propose a novel moving IRSTD network (TDCNet), which effectively extracts and enhances spatio-temporal features for accurate target detection. Specifically, we introduce a novel temporal difference convolution (TDC) re-parameterization module that comprises three parallel TDC blocks designed to capture contextual dependencies across different temporal ranges. Each TDC block fuses temporal difference and 3D convolution into a unified spatio-temporal convolution representation. This re-parameterized module can effectively capture multi-scale motion contextual features while suppressing pseudo-motion clutter in complex backgrounds, significantly improving detection performance. Moreover, we propose a TDC-guided spatio-temporal attention mechanism that performs cross-attention between the spatio-temporal features extracted from the TDC-based backbone and a parallel 3D backbone. This mechanism models their global semantic dependencies to refine the current frame’s features, thereby guiding the model to focus more accurately on critical target regions. To facilitate comprehensive evaluation, we construct a new challenging benchmark, IRSTD-UAV, consisting of 15,106 real infrared images with diverse low signal-to-clutter ratio scenarios and complex backgrounds. Extensive experiments on IRSTD-UAV and public infrared datasets demonstrate that our TDCNet achieves state-of-the-art detection performance in moving target detection. 
\end{abstract}

%Specifically, we introduce a Temporal Difference Convolution Re-parameterization (TDCR) module that directly embeds motion-sensitive filtering into the convolution process by modeling multi-scale temporal gradients, thereby enhancing target saliency while suppressing static and pseudo-motion clutter. Furthermore, a TDC-Guided Spatio-Temporal Attention (TGSTA) mechanism is incorporated into the backbone, utilizing temporal difference maps and multi-frame context to adaptively refine feature alignment and improve target-background separability. To facilitate comprehensive evaluation, we construct a new challenging benchmark, IRSTD-UAV, consisting of 15,106 frames with diverse low-SCR scenes and complex backgrounds. Extensive experiments on IRSTD-UAV and other public datasets demonstrate that TDCNet achieves state-of-the-art detection performance.

\vspace{-0.8em}
\begin{links}
\link{Code}{https://github.com/IVPLaboratory/TDCNet}
\end{links}
\vspace{-1em}

\section{Introduction}

\begin{figure}[!t]
\centering
\includegraphics[width=3.3in,keepaspectratio]{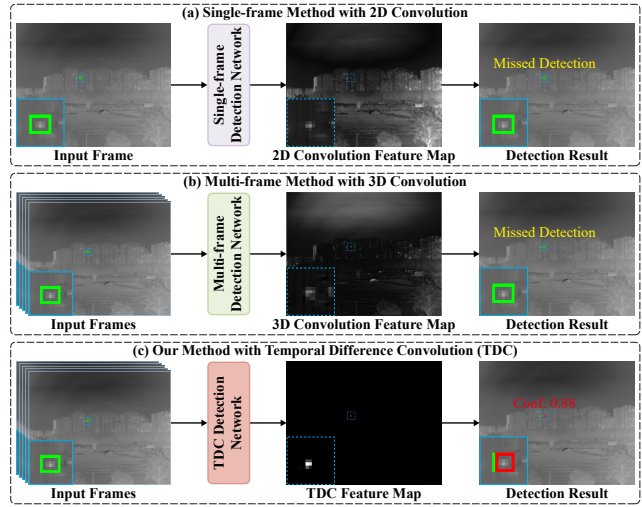} 
\caption{Three representative categories of methods for moving infrared small target detection. (a) Single-frame methods \cite{2024CVPRLiuMSHNet} employ 2D convolution, which lacks temporal context and often fails to distinguish targets from background clutter. 
(b) Multi-frame methods \cite{2025EAAIPengSTME} typically utilize 3D convolution to extract spatio-temporal features, but they often overlook explicitly leveraging motion cues, resulting in limited detection performance.
(c) Our method introduces temporal difference convolution (TDC) to explicitly capture motion-contextual information while representing spatio-temporal features, thereby effectively suppressing complex backgrounds and enhancing the detection performance of moving infrared small targets.}
\label{fig:motivation}
\end{figure}

Moving infrared small target detection (IRSTD) aims to locate small and dim targets in infrared images, often under complex backgrounds and low signal-to-clutter ratio (SCR) conditions. It plays a crucial role in a wide range of applications, including surveillance of unmanned aerial vehicles (UAVs) \cite{2022TIMFang,2023ACMMMFangDANet,2023TIIFangDAGNet,2025CVPR_UniCD,2025AAAIZhangMOCID} and space-based monitoring \cite{2021TGRSDuTempDiff}. In such scenarios, targets are typically tiny and low-contrast in the spatial domain, making them easily overwhelmed by complex dynamic background clutter. These challenges often result in missed detections and false alarms. %Given the increasing demand for robust autonomous perception in both civilian and defense domains, there is an urgent need for more effective solutions to enhance the reliability and accuracy of IRSTD in realistic, dynamic environments.

To address the above challenges, numerous infrared small target detection methods have been proposed and can be broadly divided into two categories: single-frame approaches \cite{2023ACMMMFangDANet,2023TIIFangDAGNet,2025CVPR_UniCD,2024CVPRLiuMSHNet} and multi-frame approaches \cite{2024TGRSChenSSTNet,2024TGRSTongST-Trans,2025AAAIZhangMOCID,2025EAAIPengSTME}. The former focuses on constructing complex network architectures to extract spatial features but lacks the capability to model temporal motion patterns in complex backgrounds, often leading to missed detections or false alarms \cite{2022TIMFang,2025AAAIYangPinwheelConv,2024AAAIZhangIRPruneDet,2024CVPRLiuMSHNet}. Accurate spatio-temporal feature modeling is crucial for moving IRSTD. Accordingly, the latter incorporates multi-frame inputs and leverages temporal difference modeling \cite{2021CVPRWangTDN,2023TGRSYanSTDMANet,2023CVPRXiaoLSTFENet} or spatio-temporal (3D) convolutions \cite{2025EAAIPengSTME,2025TNNLSLiDTUM} to extract spatio-temporal target features. The temporal difference operation can explicitly capture temporal contextual information but has a limited capability to extract spatial features. In contrast, 3D convolution can effectively represent features in three dimensions, yet it lacks explicit awareness of motion dynamics along the temporal dimension. This limitation hampers its capability to capture subtle frame-to-frame variations at the pixel level, which is particularly critical in detecting weak small targets under low-SCR scenarios \cite{2024TGRSHuangLMAFormer, 2025AAAIZhangMOCID,2025EAAIPengSTME, 2025TNNLSLiDTUM}. 

%sequential modeling methods
%Recent advances incorporate multi-frame inputs and adopt 3D convolution or recurrent architectures to aggregate temporal information \cite{2024TGRSChenSSTNet, 2025EAAIPengSTME, 2025TNNLSLiDTUM}. While these multi-frame methods achieve noticeable performance gains, they often treat spatial and temporal modeling as separate stages and neglect the interplay between them. As a result, they fail to fully exploit motion cues that are crucial for enhancing target saliency and suppressing background interference.

%Most existing multi-frame IRSTD methods suffer from three major limitations. First, existing multi-frame IRSTD methods often adopt complex temporal architectures but lack the ability to directly encode fine-grained inter-frame motion differences, limiting their effectiveness in low-SCR and cluttered environments \cite{2024TGRSHuangLMAFormer, 2025AAAIZhangMOCID}. Second, general 3D convolutions are limited in their ability to represent subtle frame-to-frame variations at pixel level, which are particularly important in detecting weak and fast-moving targets \cite{2025EAAIPengSTME, 2025TNNLSLiDTUM}. Third, current attention-based methods primarily focus on semantic-level dependencies across frames, but lack the ability to align and enhance motion-sensitive features \cite{2024TGRSTongST-Trans, 2025IVCZhuSTC}. These gaps result in feature contamination, poor target-background separability, and frequent missed detections in cluttered scenes.

To overcome the above limitations, we propose a novel moving IRSTD network (TDCNet), which effectively extracts and enhances spatio-temporal features for accurate target detection. In general, infrared small targets have weak features in the spatial domain and are susceptible to complex background interference. However, moving infrared small targets typically exhibit strong motion-contextual dependencies along the temporal dimension. This observation motivates us to exploit such contextual dependencies to suppress complex background interference and more effectively model the spatio-temporal features. In this work, we introduce the temporal difference convolution re-parameterization (TDCR) module that integrates three parallel temporal difference convolution (TDC) blocks to model short‑, mid‑, and long‑term motion-contextual dependencies, respectively. Each TDC block fuses temporal difference and 3D convolution into a unified spatio-temporal convolution representation, which is designed to effectively capture motion-contextual dependencies within a specified temporal range. This allows the TDC block to suppress background clutter as it simultaneously enhances the discrimination of spatio-temporal features for infrared small targets. The multi-branch architecture of the TDCR module during training is equivalently converted into a single-branch structure for efficient inference. This design enables the TDCR module to effectively capture multi-scale motion-contextual dependencies while suppressing pseudo-motion clutter in complex backgrounds without incurring additional computational cost during inference, significantly improving detection performance. 

Moreover, we propose a TDC-guided spatio-temporal attention mechanism that performs cross-attention between the spatio-temporal features extracted from the TDC-based backbone and a parallel 3D convolution backbone. The TDC-based backbone emphasizes the salient target regions while suppressing interference from complex backgrounds. By leveraging this property, the proposed mechanism effectively captures global semantic dependencies between the two feature streams and refines the spatio-temporal feature representation of the current frame, guiding the model to focus more accurately on critical target regions and thereby enhancing detection performance. Furthermore, we construct a new IRSTD benchmark, termed IRSTD-UAV, which contains 15,106 frames captured across diverse UAV types and complex backgrounds. Extensive experiments on both IRSTD-UAV and public benchmark IRDST \cite{2023IRDST} demonstrate that TDCNet achieves state-of-the-art (SOTA) detection performance, significantly outperforming existing single-frame and multi-frame methods under low SCR and complex backgrounds. 

The contributions of this work can be summarized as:
\begin{itemize}
\item We introduce a novel moving IRSTD network (TDCNet) that can effectively capture spatio-temporal features while suppressing complex backgrounds for accurate detection.
\item We are the first to propose TDC that fuses temporal difference and spatio-temporal convolution into a unified 3D convolution representation, enabling effectively capture motion-contextual dependencies within a specified temporal range. 
\item We introduce a novel TDC-guided spatio-temporal attention (TDCSTA) mechanism that models semantic relationships between TDC-enhanced features and parallel 3D convolutional features. This mechanism is leveraged to refine the representations of critical target regions in the current frame, thereby enhancing the detection performance in complex backgrounds.
\end{itemize}

\section{Related Work}

\begin{figure*}[!t]
\centering
\includegraphics[width=7in,keepaspectratio]{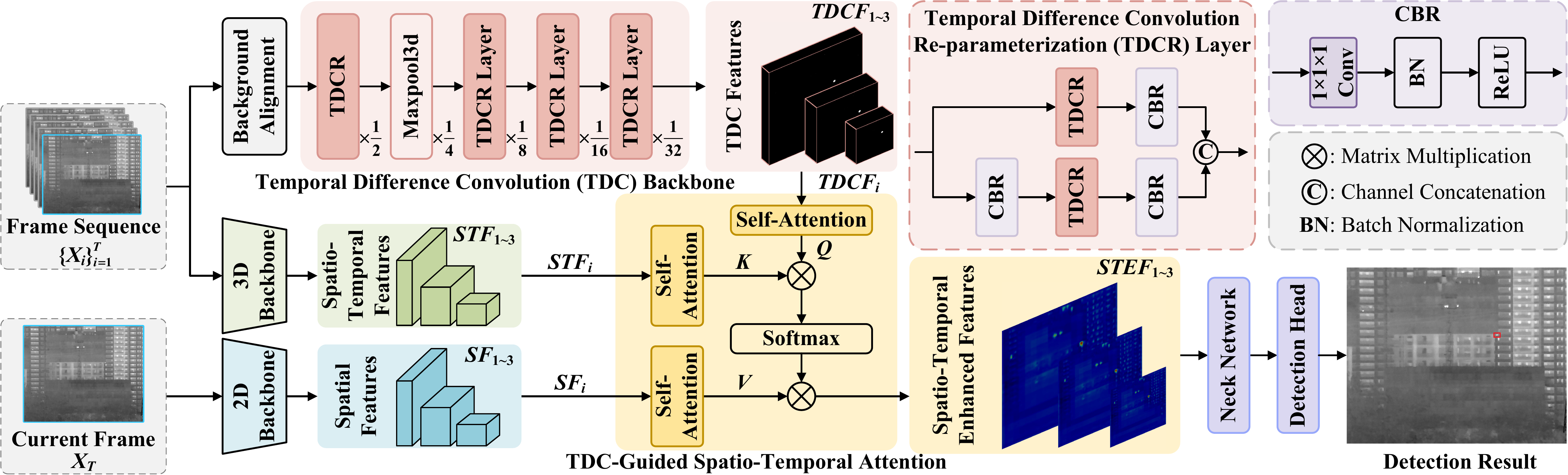} 
\caption{Overview of the proposed TDCNet. The input consists of a frame sequence $\{X_i\}_{i=1}^T$ and the current frame $X_T$. The temporal difference convolution (TDC) backbone utilizes the temporal difference convolution re-parameterization layer to extract TDC features from $\{X_i\}_{i=1}^T$. The 2D backbone processes $X_T$ to extract spatial features, while the 3D backbone handles $\{X_i\}_{i=1}^T$ to extract spatio-temporal features. The TDC-guided spatio-temporal attention module refines these feature streams to generate spatio-temporal enhanced features, which are aggregated by the neck and detection head to produce the final detection result.}
\label{fig:overall}
\end{figure*}

% \subsection{Single-Frame Infrared Small Target Detection}
% Traditional single-frame infrared small target detection methods primarily focus on enhancing spatial feature representation to distinguish small targets from complex backgrounds. For example, DAGNet \cite{2023TIIFangDAGNet} and DANet \cite{2023ACMMMFangDANet} incorporate attention mechanisms to improve the network’s ability to extract and refine target features from complex backgrounds. ISNet \cite{2022CVPRZhangISNet} enhances shape-related features via Taylor finite difference-based edge modeling and orientation-aware attention, improving robustness under heavy clutter. MSHNet \cite{2024CVPRLiuMSHNet} addresses loss function limitations by proposing a scale- and location-sensitive (SLS) loss that significantly boosts detection performance, even with a lightweight backbone. Despite these advances, single-frame approaches inherently lack the capacity to exploit temporal motion cues critical for disambiguating small targets from dynamic or pseudo-motion backgrounds, motivating the integration of temporal information in multi-frame frameworks.

\subsection{Moving Infrared Small Target Detection}
Existing moving IRSTD methods mainly differing in how they handle spatial and temporal information. A widely adopted strategy is to apply 2D convolutional networks independently on each frame within a temporal sequence~\cite{2023TGRSYanSTDMANet,2024TGRSChenSSTNet}. However, the lack of inter-frame interaction constrains their capability to model spatio-temporal continuity. Conversely, temporal difference methods focus exclusively on frame-wise intensity variations to capture motion cues~\cite{2021TGRSDuTempDiff,2023TGRSYanSTDMANet}, but struggle to extract the spatial semantic representations essential for robust detection. To effectively leverage both spatial and temporal information, recent approaches either adopt staged pipelines that first extract spatial features via 2D convolutions and then apply temporal modeling~\cite{2025AAAIZhangMOCID, 2025IVCZhuSTC} or employ 3D convolutions to jointly capture spatio-temporal features~\cite{2025JSTARLSTD,2025TNNLSLiDTUM}. However, these methods often suffer from limited motion-awareness or insufficient spatio-temporal context modeling in complex backgrounds. In contrast, we propose TDC block that fuses temporal difference and spatio-temporal convolution into a unified 3D convolution representation, effectively capturing motion-contextual dependencies for robust moving IRSTD under complex backgrounds.

\subsection{Spatio-Temporal Context Modeling}
Temporal difference, 3D convolution, and transformer-based models are fundamental techniques for spatio-temporal modeling in video analysis, widely applied to tasks such as action recognition and video understanding~\cite{2018NIPS_TC4ActRec, 2021CVPRWangTDN, bertasius2021space}. Temporal difference captures inter-frame variations to highlight motion cues~\cite{2018WACVTDN,2023TMMGTDNet}, while 3D convolution jointly learns spatial and temporal features~\cite{2018CVPRMiCT,li2019ACMMMDCTVonv}. Transformer-based models further introduce temporal self-attention to enable long-range dependency modeling~\cite{2021CVPRViViT,2023CVPRVT}. However, each method focuses on a limited aspect: temporal difference lacks semantic context, and both 3D convolution and transformer-based models often overlook explicit motion cues. In this work, we propose a unified spatio-temporal network that combines multi-scale motion-contextual modeling via TDCR and spatio-temporal feature enhancement via TDCSTA for robust moving IRSTD.

\section{The Proposed Method}
\subsection{Overall Architecture}
% In this section, we propose a novel multi-frame detection network, TDCNet, as illustrated in Figure \ref{fig:overall}. TDCNet first employs a temporal difference backbone to extract motion-sensitive features by computing multi-scale temporal gradients. Then, a TDC-guided spatio-temporal attention module adaptively fuses and aligns spatial, temporal, and temporal difference features through a unified attention mechanism, thereby enhancing motion-aware representations. Finally, we construct a new benchmark dataset, IRSTD-UAV, to comprehensively validate the effectiveness of our approach.

In this study, we propose a novel moving IRSTD network, TDCNet, as illustrated in Figure~\ref{fig:overall}. It first introduces a temporal difference convolution (TDC) backbone. Then, a TDC-guided spatio-temporal attention module refines feature representations by applying self-attention to three distinct feature streams and performing cross-attention with TDC features as the query to selectively enhance spatio-temporal features. Finally, we construct a challenging benchmark dataset, IRSTD-UAV, to validate the effectiveness of our method.

\subsection{Temporal Difference Convolution Backbone}
Inspired by the 3D backbone design of STMENet~\cite{2025EAAIPengSTME}, the TDC backbone is introduced to extract spatio-temporal contextual features. Before the frame sequence is fed into the TDC backbone, a background alignment process is applied to suppress camera motion~\cite{shen2024gimlearninggeneralizableimage}. By progressively stacking TDCR layers, the spatio-temporal contextual features from earlier stages are further refined with multi-scale temporal ranges, enabling the model to learn more discriminative representations of small moving targets embedded in complex infrared scenes.%The TDC backbone builds upon the 3D backbone used in STMENet~\cite{2025EAAIPengSTME}, where each 3D convolution is replaced with the proposed temporal difference convolution re-parameterization (TDCR) module, forming the TDCR Layers. 

\subsection{Temporal Difference Convolution Re-parameterization Module}
\begin{figure}[!t]
\centering
\includegraphics[width=3.3in,keepaspectratio]{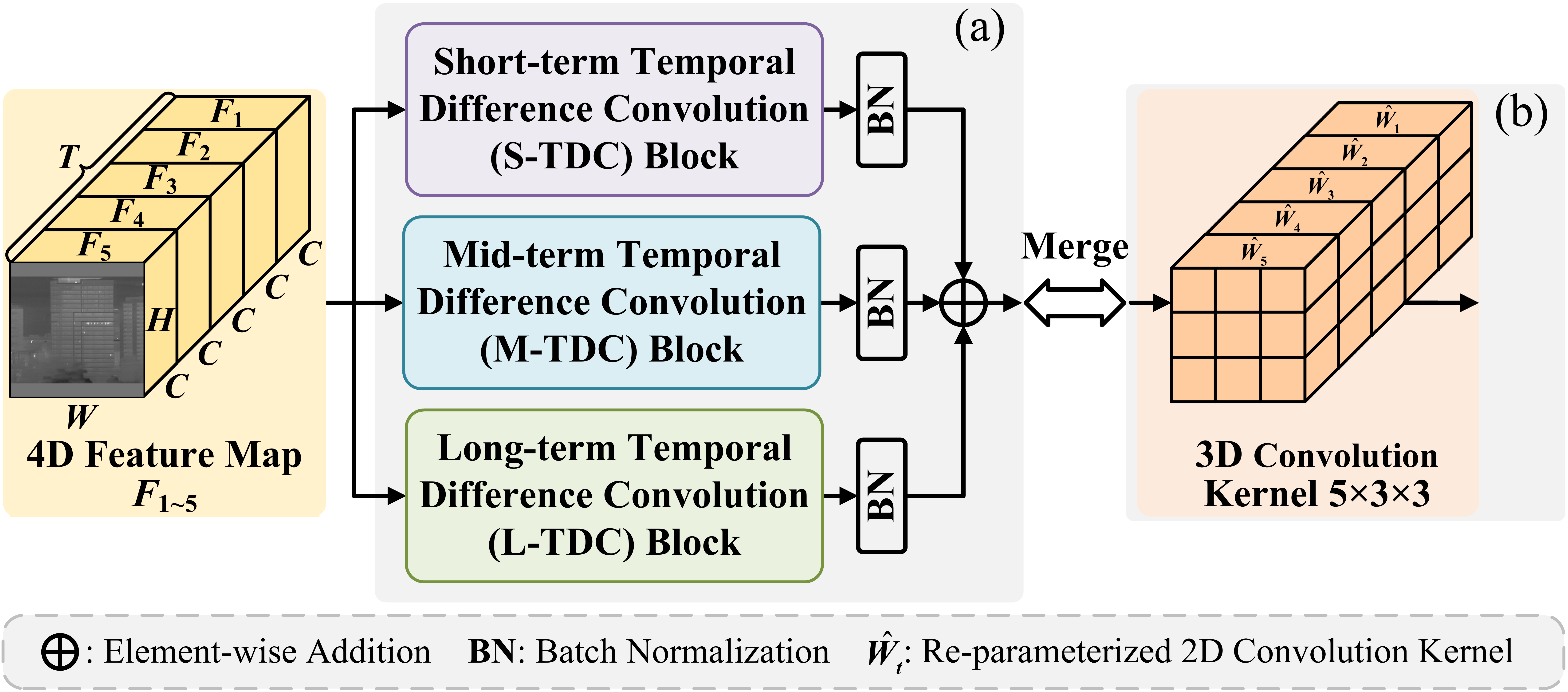} \vspace{-1.5em}
\caption{Overview of the proposed temporal difference convolution re-parameterization (TDCR) module, which equivalently transforms three parallel TDC blocks (a) into a single 3D convolution representation (b).}
\label{fig:tdcr}
\end{figure}

\begin{figure}[!t]
\centering
\includegraphics[width=3.3in,keepaspectratio]{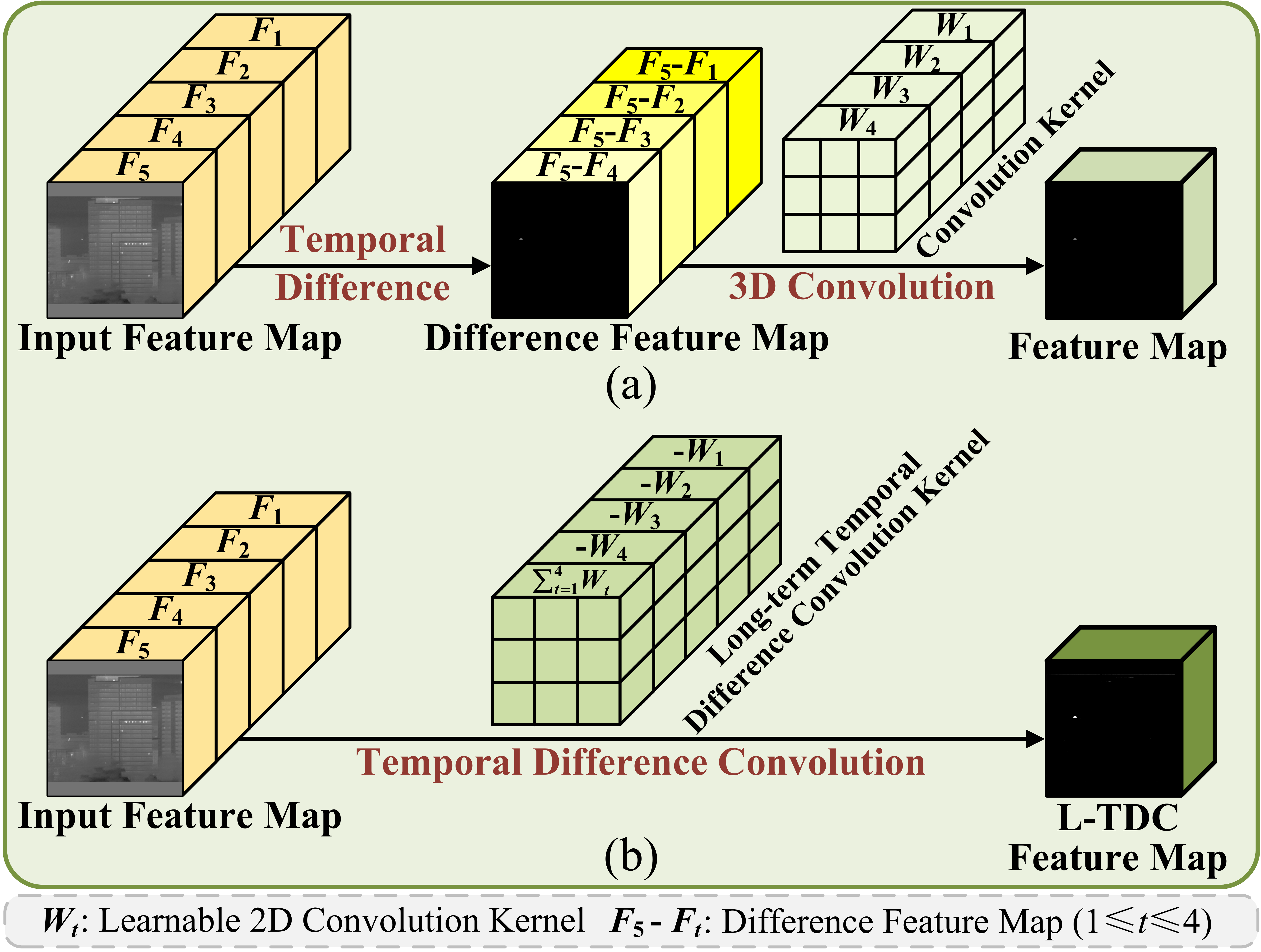} \vspace{-1.5em}
\caption{(a) Sequential combination of existing temporal difference and 3D convolution, and (b) Our proposed long-term temporal difference convolution (L-TDC) block, which fuses temporal difference modeling and 3D convolution into a unified spatio-temporal convolution representation. }
\label{fig:ltdc}
\end{figure}

As illustrated in Figure~\ref{fig:tdcr}, we propose a novel TDCR module to enhance spatio-temporal contextual feature modeling capability over multiple temporal scales. During training, the TDCR consists of three parallel branches: short-term TDC (S-TDC) block, mid-term TDC (M-TDC) block, and long-term TDC (L-TDC) block (Figure~\ref{fig:ltdc}). Each branch is specifically designed to capture temporal dependencies at different temporal scales. The outputs of these blocks are independently normalized by batch normalization layers and then aggregated through summation. During inference, we re-parameterize the three branches into a unified single 3D convolution to simplify the inference pipeline while preserving the multi-scale temporal modeling capability.

\subsubsection{Temporal Difference Convolution.}
Accurate spatio-temporal feature modeling is essential for robust moving IRSTD in infrared sequences. Traditional approaches typically rely on either temporal difference operations or 3D convolutions. The temporal difference directly models the motion information by computing frame-wise differences, providing strong awareness of motion dynamics but suffering from weak spatial feature representation \cite{2021TGRSDuTempDiff}. In contrast, 3D convolution effectively extracts spatio-temporal features but lacks explicit motion awareness in cluttered backgrounds \cite{2025EAAIPengSTME}. To leverage the strengths of both methods, we propose the TDC block, which fuses temporal difference and 3D convolution into a unified spatio-temporal convolution representation. Specifically, to explicitly capture the motion-contextual dependencies between frames, we reformulate the traditional 3D convolutional weights \( W \in \mathbb{R}^{C_{\text{out}} \times C_{\text{in}} \times T \times H \times W} \), where \( C_{\text{in}} \) and \( C_{\text{out}} \) denote the number of input and output channels, respectively, and \( T \), \( H \), and \( W \) represent the kernel size along the temporal, height, and width dimensions. As presented in Figure~\ref{fig:ltdc}, we take the L-TDC block as an example. The input feature map $F \in \mathbb{R}^{T \times C \times H \times W}$ to the L-TDC block consists of a sequence of frames $\{F_t\}_{t=1}^{5}$, where each $F_t \in \mathbb{R}^{C \times H \times W}$. Here, $F_5$ denotes the current frame, while $F_1$ to $F_4$ are preceding frames. The L-TDC block aims to capture long-term motion-contextual dependencies by computing differences between the current frame and all previous frames. To achieve this, $W$ is decomposed along the temporal dimension into a set of 2D convolution kernels $\{W_t\}_{t=1}^{4}$, where each $W_t \in \mathbb{R}^{C_{\text{out}} \times C_{\text{in}} \times H \times W}$ models the frame-wise difference at time step $t$. Its output is defined as:
\begin{align} \label{eq:ltdc}
F_l&=(-W_{1})*F_{1}+(-W_{2})*F_{2}+(-W_{3})*F_{3}\nonumber\\
&\quad +(-W_{4})*F_{4}+ \left(\textstyle \sum_{t=1}^{4}W_{t} \right) *F_{5}\nonumber\\
&=\textstyle \sum_{t=1}^{4}\left(W_{t} *(F_{5}-F_{t})\right).
\end{align}  
Here, $\ast$ denotes the convolution operation. $F_l$ is mathematically equivalent to the summation of convolutions between $W_t$ and the temporal difference feature map $F_5-F_t$. However, as shown in Figure \ref{fig:ltdc}, it is important to emphasize that our TDC does not explicitly perform a difference operation. Instead, it implicitly fuses temporal difference and 3D convolution into a unified spatio-temporal convolution representation. This formulation explicitly encodes the long-term temporal difference and rich spatio-temporal features between the current frame and all previous frames, capturing long-term motion-contextual dependencies. 

S-TDC and M-TDC are derived similarly to L-TDC, each targeting motion modeling at different temporal scales. The S-TDC block focuses on short-term motion by computing differences between consecutive frames:
\begin{align} \label{eq:stdc}
F_s&=(-W_{2})*F_{1}+(W_{2}-W_{3})*F_{2}+(W_{3}-W_{4})*F_{3}\nonumber\\
&\quad +(W_{4}-W_{5})*F_{4}+W_{5}*F_{5}\nonumber \\
&=\textstyle \sum_{t=2}^{5}\left(W_{t} *(F_{t}-F_{t-1})\right).
\end{align}
This short-term motion modeling design enhances the network’s sensitivity to fine-grained and fast-changing motion patterns, enabling effective capture of subtle variations between consecutive frames. Meanwhile, the M-TDC block captures intermediate motion context by computing differences between frames with two-frame intervals, complementing short- and long-term modeling with its distinct temporal scope:
\begin{align} \label{eq:mtdc}
F_m&=(-W_{3})*F_{1}+(-W_{4})*F_{2}+(W_{3}-W_{5})*F_{3}\nonumber\\
&\quad +W_{4}*F_{4}+W_{5}*F_{5} \nonumber\\
&=\textstyle \sum_{t=3}^{5}\left(W_{t}*(F_{t}-F_{t-2})\right).
\end{align}
This design enables the network to capture mid-term motion contextual dependencies while mitigating redundant motion or noise. Together, the three TDC blocks capture complementary spatio-temporal features at different temporal scales, thereby strengthening the overall motion context modeling capability.

As a result, the TDCR module captures multi-scale motion-contextual dependencies through three parallel TDC branches: $\tilde{F}_s = \text{BN}_s(F_s), \quad \tilde{F}_m = \text{BN}_m(F_m), \quad \tilde{F}_l = \text{BN}_l(F_l)$, where $\text{BN}_{\{s,m,l\}}$ are their corresponding batch normalization layers. The three outputs are then aggregated as the final output of the TDCR module: $F_{TDCR} = \tilde{F}_s + \tilde{F}_m + \tilde{F}_l$.

\subsubsection{Re-parameterization of the Multi-scale TDC Branches.}
% \subsubsection{Re-parameterization during Inference.}
We first fuse convolution and BN operations within each TDC branch via parameter transformation~\cite{2024WACVkobayashi3DCNN}: 
$
\hat{W}_i = \gamma_i \cdot W_i / \sigma_i, \quad
\hat{b}_i = \gamma_i \cdot (b_i - \mu_i) / \sigma_i + \beta_i,
$
where \( W_i \) and \( b_i \) denote the convolutional kernel weights and biases, and \( \gamma_i, \beta_i, \mu_i, \sigma_i \) are the parameters of batch normalization. Leveraging the homogeneity and additivity of convolution, we then merge the three TDC branches into a single 3D convolution:
$
\hat{W}_{\text{TDCR}} = \sum_{i \in \{s,m,l\}} \hat{W}_i, \quad
\hat{b}_{\text{TDCR}} = \sum_{i \in \{s,m,l\}} \hat{b}_i.
$
The resulting re-parameterized TDCR module can be expressed as:
$
\text{TDCR}(F) = \hat{W}_{\text{TDCR}} \ast F + \hat{b}_{\text{TDCR}}.
$
This re-parameterization improves intra-model inference efficiency while preserving the benefits of multi-scale motion context modeling.
%Here, “reduction in computational cost” refers to the intra-model efficiency gain achieved after re-parameterization, rather than a cross-method comparison.

\subsection{TDC-Guided Spatio-Temporal Attention Module}

As illustrated in the bottom center of Figure~\ref{fig:overall}, we propose a TDC-guided spatio-temporal attention (TDCSTA) module to refine the feature representation of small moving targets in cluttered infrared scenes. Unlike conventional methods that directly fuse multi-frame features, TDCSTA introduces a tri-branch architecture to decouple and specialize in distinct spatio-temporal cues, thereby enabling more structured and effective feature interaction. Specifically, TDCSTA operates on three feature streams extracted from the final three stages of their respective backbones: the temporal difference convolution features ($TDCF_{1\sim3}$) from the TDC backbone, the spatio-temporal features ($STF_{1\sim3}$) from the 3D backbone and the spatial features ($SF_{1\sim3}$) from the 2D backbone. By capturing global semantic dependencies and enabling selective feature interaction, TDCSTA generates spatio-temporal enhanced features, helping the model focus more accurately on small targets and improving detection performance. The enhanced features are then passed to the neck and detection head to produce the final detection result.

% At each level $i$, the corresponding features $\text{TDCF}_i$, $\text{STF}_i$, $\text{SF}_i$ are aligned in spatial resolution and channel dimensionality.

\subsubsection{Self-Attention for Semantic Expressiveness Enhancement.}
To enhance the semantic representation capability of each feature stream and effectively suppress irrelevant background clutter, we apply a self-attention mechanism independently to the three feature streams: $TDCF_i$, $STF_i$, and $SF_i$, at each stage $i$. We partition each feature stream $FS \in \mathbb{R}^{T \times C \times H \times W}$ into non-overlapping 3D local windows of size $P \times M \times M$, and compute self-attention with both regular and shifted window partitioning applied~\cite{liu2021videoswintransformer}. Formally, the self-attention is defined as:
\begin{equation}
\text{SA}(FS) = \text{Softmax}( Q K^\top / \sqrt{d} + B ) V,
\end{equation}
where $Q, K, V \in \mathbb{R}^{PM^2 \times d}$ are linear projections of the input tokens within each window, $d$ is the embedding dimension and $B$ is the relative positional bias. We apply this mechanism to each feature stream as follows:$\hat{F}_{TDCF,i} = \text{SA}(TDCF_i),\quad
\hat{F}_{STF,i} = \text{SA}(STF_i),\quad
\hat{F}_{SF,i} = \text{SA}(SF_i).$

\subsubsection{Cross-Attention for TDC-Guided Semantic Dependency Modeling.}
To explicitly model semantic dependencies guided by motion-aware features, we employ a cross-attention mechanism where $\hat{F}_{TDCF,i}$ serves as the query, while $\hat{F}_{STF,i}$ and $\hat{F}_{SF,i}$ serve as key and value, respectively. The output of the cross-attention mechanism is the spatio-temporal enhanced features (STEF), defined as:
\begin{equation}
STEF_{i} = \text{Softmax} ( Q_i K_i^\top / \sqrt{d} + B_i ) V_i,
\end{equation}
where $Q_i$ is derived from $\hat{F}_{TDCF,i}$, $K_i$ from $\hat{F}_{STF,i}$, and $V_i$ from $\hat{F}_{SF,i}$.
By leveraging the discriminative motion-contextual cues encoded in $\hat{F}_{TDCF,i}$, which highlight salient target regions while suppressing complex background interference, this mechanism enables the model to focus on semantically relevant regions across temporal and spatial dimensions, thereby enhancing semantic dependency modeling and refining the spatio-temporal representation of the current frame.

% The $TDCF$ highlights salient target regions while suppressing complex background interference. Leveraging this property, the cross-attention effectively models global semantic dependencies across both feature streams, refining the spatio-temporal representation of the current frame. This guided interaction enables the model to focus more precisely on critical target regions, thereby improving detection accuracy.

\section{Experimental Results and Analysis}
\subsection{Datasets and Evaluation Metrics}
\subsubsection{Datasets.}
We evaluate our method on two real infrared benchmarks: a self-constructed IRSTD-UAV dataset and a public IRDST dataset \cite{2023IRDST}. The IRSTD-UAV dataset contains 17 real infrared video sequences with 15,106 frames, featuring small targets and complex backgrounds such as buildings, trees, and clouds. More details of our dataset are provided in the supplementary material.

\subsubsection{Evaluation Metrics.}
For evaluation, we adopt standard metrics, including precision (P), recall (R), F\textsubscript{1}-score (F\textsubscript{1}), and average precision (AP\textsubscript{50}), all computed at an intersection-over-union (IoU) threshold of 0.5. Real-time performance is measured in frames per second (FPS), while computational complexity is assessed using the number of parameters (Params) and floating point of operations (FLOPs).

\begin{table*}[!t]
\centering
\fontsize{8.5}{10}\selectfont %font size and line height
\setlength\tabcolsep{3pt}
\begin{tabular}{c|cc|cccc|cccc|p{4.5em}<{\centering}p{4em}<{\centering} p{2em}<{\centering}} %p{2em}<{\centering}局部调整列间距
\hline
\multirow{2}{*}{Type} & \multirow{2}{*}{Method} & \multirow{2}{*}{Pub'Year} & \multicolumn{4}{c|}{IRSTD-UAV} & \multicolumn{4}{c|}{IRDST} & \multirow{2}{*}{Params (M)} & \multirow{2}{*}{FLOPs (G)} & \multirow{2}{*}{FPS} \\ \cline{4-7} \cline{8-11}
 &  &  & P & R & $\text{F}_1$ & $\text{AP}_{50}$ & P & R & $\text{F}_1$ & $\text{AP}_{50}$ &  &  &\\ \hline

\multirow{5}{*}{\makecell[c]{Single-\\frame}} 
 & YOLO11-L & 2024 & 96.26 & \underline{95.73} & 95.99 & 91.20 & 96.70 & 96.00 & 96.35 & 92.10 & 26.0 & 111.8 & 61.7 \\
 & MSHNet & CVPR'2024 & 86.92 & 84.94 & 85.92 & 87.62 & 82.31 & 77.64 & 79.91 & 63.21 & 4.1 & 76.3 & 63.3 \\ 
 & SCTransNet & TGRS'2024 & 96.71 & 89.19 & 92.80 & 85.36 & 96.80 & 90.18 & 93.37 & 92.35 & 11.2 & 80.9 & 17.8 \\
 & YOLOv12-L & 2025 & 96.50 & 94.99 & 95.74 & 90.54 & 97.29 & 95.63 & 96.45 & 92.09 & 27.1 & 104.8 & 30.5 \\
 & PConv (YOLOv8) & AAAI'2025 & 94.51 & 94.00 & 94.25 & 88.47 & 96.21 & 95.91 & 96.06 & 91.88 & 23.8 & 88.6 & 58.8 \\
 & Hyper-YOLO-M & TPAMI'2025 & 96.21 & 95.08 & 95.64 & 91.04 & 96.68 & \underline{96.90} & 96.79 & 92.53 & 32.4 & 119.0 & \textbf{87.3} \\ \hline

\multirow{8}{*}{\makecell[c]{Multi-\\frame}} 
 & TMP & ESA'2024 & 37.90 & 61.70 & 46.96 & 23.00 & 86.65 & 81.36 & 83.92 & 70.01 & 16.4 & 92.9 & 25.0 \\
 & SSTNet & TGRS'2024 & 88.98 & 84.19 & 86.52 & 74.60 & 88.56 & 81.92 & 85.11 & 71.55 & 11.9 & 123.6 & 22.6 \\
 & MOCID & AAAI'2025 & \underline{97.28} & 94.85 & \underline{96.05} & \underline{91.32} & \textbf{98.92} & 96.86 & \underline{97.88} & \underline{94.74} & 13.1 & 98.7 & 11.5 \\
 & STMENet & EAAI'2025 & 86.70 & 88.04 & 87.36 & 75.97 & 87.78 & 84.22 & 85.96 & 73.40 & 10.4 & \underline{45.8} & 48.5 \\
 & ResUNet+RFR & TGRS'2025 & 56.24 & 82.47 & 66.87 & 46.20 & 83.92 & 76.10 & 79.82 & 61.45 & \underline{0.9} & 73.8 & \underline{72.6} \\
 & ResUNet+DTUM & TNNLS'2025 & 93.83 & 83.17 & 88.18 & 81.37 & 82.87 & 87.79 & 85.26 & 71.48 & \textbf{0.3} & \textbf{41.4} & 13.9 \\
 & \textbf{Ours} & - & \textbf{97.99} & \textbf{96.27} & \textbf{97.12} & \textbf{93.83} & \underline{98.12} & \textbf{97.71} & \textbf{97.91} & \textbf{94.79} & 24.8 & 95.7 & 18.5 \\ \hline
\end{tabular} \vspace{-0.5em}
\caption{Quantitative comparison of the proposed method with SOTA methods on the IRSTD-UAV and IRDST datasets. \textbf{Bold} and \underline{underline} indicate the best and the second best results, respectively.}
\label{table1}
\end{table*}

\begin{figure*}[!t]
\centering
\includegraphics[width=6.7in,keepaspectratio]{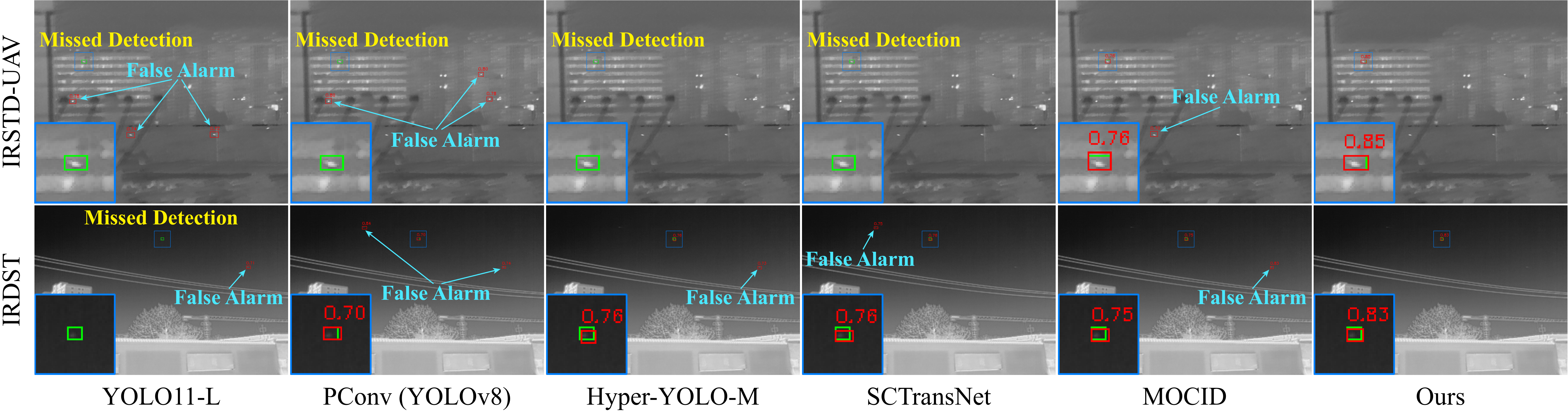} \vspace{-0.5em}
\caption{Visual comparison of results from SOTA methods and TDCNet on the IRSTD-UAV and IRDST dataset. Boxes in green and red represent ground-truth and detected targets, respectively.}
\label{fig:vis}
\end{figure*}

\subsection{Implementation Details}
All experiments are conducted on a single NVIDIA RTX 3090 GPU with CUDA 12.4 and PyTorch 2.7. The training is performed using the Adam optimizer with a learning rate of 0.001 and a weight decay of $1 \times 10^{-4}$. We first pre-train the 2D backbone on still images and the 3D backbone on multi-frame inputs. Both are then frozen, and the TDC backbone along with the TDCSTA module is trained on video sequences. The input frames are resized to $640 \times 640$, and five consecutive frames were used as input during training and inference. We adopt the IoU loss for regression and binary cross-entropy loss for objectness and classification: \(\mathcal{L} = \mathcal{L}_{reg} + \mathcal{L}_{obj} + \mathcal{L}_{cls}\). For single-frame methods, we include YOLO11-L \cite{Jocher_Ultralytics_YOLO11}, YOLOv12-L \cite{tian2025yolov12}, and Hyper-YOLO-M \cite{feng2024hyperyolo} as general-purpose convolutional neural network (CNN)-based detectors, along with MSHNet \cite{2024CVPRLiuMSHNet} and PConv (YOLOv8) \cite{2025AAAIYangPinwheelConv}, which are specifically designed for IRSTD. Additionally, we include SCTransNet \cite{2024TGRSYuanSCTransNet} as a infrared-specific transformer-based baseline. For multi-frame methods, we select infrared-specific CNN-based methods including TMP \cite{2024ZhuESATMP}, SSTNet \cite{2024TGRSChenSSTNet}, MOCID \cite{2025AAAIZhangMOCID}, STMENet \cite{2025EAAIPengSTME}, RFR \cite{Ying_2025RFR}, and DTUM \cite{2025TNNLSLiDTUM}.

\subsection{Quantitative Results}

As shown in Table~\ref{table1}, our proposed TDCNet achieves SOTA performance on P, R, F\textsubscript{1}, and AP\textsubscript{50} on the IRSTD-UAV dataset, and on R, F\textsubscript{1}, and AP\textsubscript{50} on IRDST. TDCNet outperforms all single-frame methods such as MSHNet and Hyper-YOLO-M, which suffer from limited robustness in cluttered infrared scenes due to the absence of temporal modeling. Among multi-frame methods, TDCNet achieves the highest R, F\textsubscript{1}, and AP\textsubscript{50}. Other methods like MOCID and SCTransNet are less effective in complex scenes due to insufficient motion modeling and suboptimal spatio-temporal representation. TDCNet achieves a lower computational cost of 95.7G FLOPs and a reasonable inference speed of 18.5 FPS.

\subsection{Qualitative Results}
As demonstrated in Figure~\ref{fig:vis}, our TDCNet exhibits superior detection performance across two challenging infrared scenarios from the IRSTD-UAV and IRDST datasets. Even in the presence of strong background clutter, such as urban structures and light-like distractors, TDCNet effectively highlights true UAV targets while suppressing false alarms. This is because the TDCR module can effectively capture multi-scale motion-contextual dependencies, while the TDCSTA selectively enhances target-relevant features while suppressing irrelevant background clutter. YOLO11-L, Hyper-YOLO-M, and PConv (YOLOv8) all struggle under complex infrared scenes, frequently missing true targets and generating false alarms due to their lack of temporal modeling and motion-aware feature representation. SCTransNet suffers from false alarms due to the absence of explicit motion context guidance. Although MOCID incorporates motion context, it fails to capture multi-scale temporal dependencies, which limits its capability to suppress clutter in complex backgrounds. More visual results are provided in the supplementary material.

% \begin{figure}[!t]
% \centering
% \includegraphics[width=2.6in,keepaspectratio]{all_methods_pr_curve.pdf} 
% \caption{P-R curves of our method and other IRSTD methods on IRSTD-UAV. The area values under the
% curves are placed after the method names.}
% \label{fig:pr_curves}
% \end{figure}

\subsection{Ablation Study}
In this section, we report ablation studies. More experiments are provided in the supplementary material.

\subsubsection{Impact of the proposed TDCR and TDCSTA.}

\begin{figure}[!t]
\centering
\includegraphics[width=3.3in,keepaspectratio]{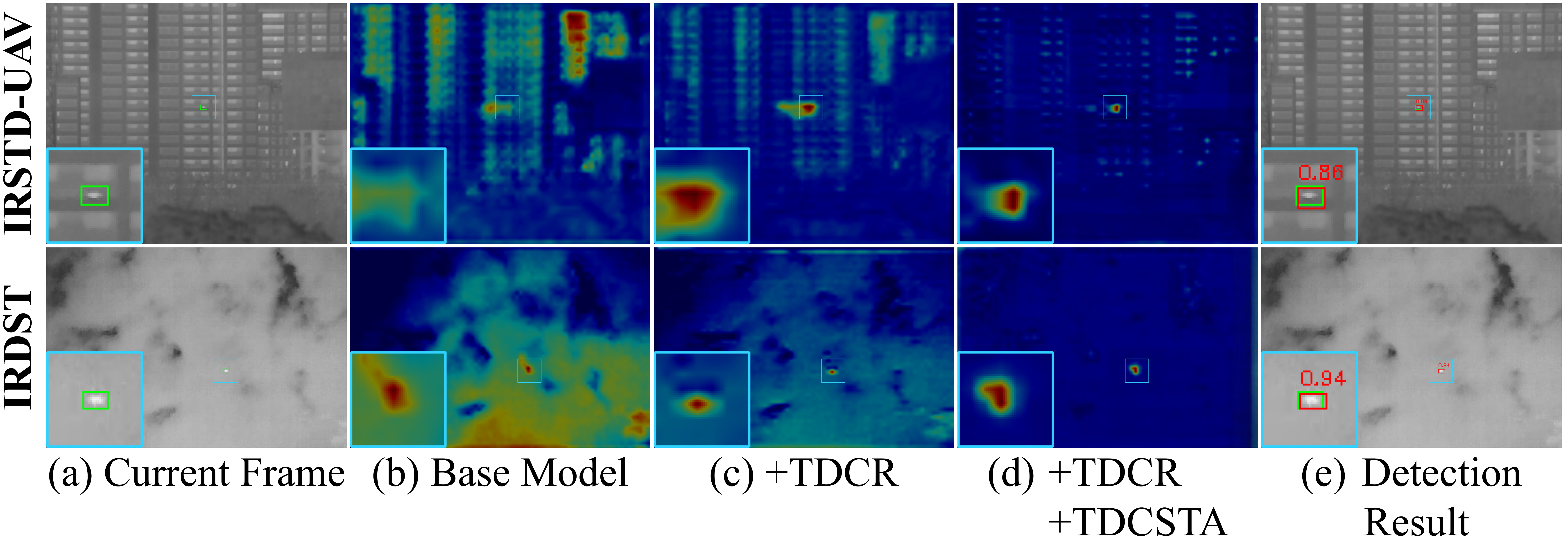} \vspace{-1.8em}
\caption{Heatmap visualization illustrating the progressive enhancement in background suppression achieved by TDCR and TDCSTA.}
\label{fig:heatmaps}
\end{figure}

As shown in Table~\ref{tab:module_ablation}, both TDCR and TDCSTA independently contribute to performance improvements over the baseline. Specifically, TDCR improves P to 97.61 and AP\textsubscript{50} to 92.50, while TDCSTA improves R to 95.96 and F\textsubscript{1} to 96.74. In combination, TDCR and TDCSTA achieve superior results, highlighting their complementary benefits. To better understand their effects, we visualize the heatmaps in Figure~\ref{fig:heatmaps}. Compared to the base model, TDCR yields more focused and distinguishable activations on targets. After applying TDCSTA, irrelevant background activations are noticeably suppressed, further enhancing target saliency in cluttered infrared scenes.

\begin{table}[!t]
\centering
\fontsize{9}{10}\selectfont %font size and line height
\setlength{\tabcolsep}{5pt}
\begin{tabular}{cc|cccc}
\hline
TDCR & TDCSTA & P & R & $\text{F}_1$ & $\text{AP}_{50}$ \\ \hline
$\times$ & $\times$ & 86.70 & 88.04 & 87.36 & 75.97 \\ 
$\surd$ & $\times$ & \underline{97.61} & 95.93 & \underline{96.76} & \underline{92.50} \\
$\times$ & $\surd$ & 97.53 & \underline{95.96} & 96.74 & 92.35 \\
$\surd$ & $\surd$ & \textbf{97.99} & \textbf{96.27} & \textbf{97.12} & \textbf{93.83} \\ \hline
\end{tabular}\vspace{-0.5em}
\caption{Ablation study of the TDCR and TDCSTA modules.}
\label{tab:module_ablation}
\end{table}

\begin{table}[!t]
\centering
\fontsize{9}{10}\selectfont %font size and line height
\setlength\tabcolsep{3pt}
\begin{tabular}{cccccc}
\hline
Method & P & R & $\text{F}_1$ & $\text{AP}_{50}$ & FLOPs (G) \\ \hline
TD & 94.36 & 90.24 & 92.25 & 89.73 & \textbf{41.351} \\ 
3D Conv & 86.70 & 88.04 & 87.36 & 75.97 & 45.842 \\
TD + 3D Conv & \underline{95.15} & \underline{92.62} & \underline{93.87} & \underline{89.81} & 45.854 \\
TDC & \textbf{97.61} & \textbf{95.93} & \textbf{96.76} & \textbf{92.50} & \underline{45.749}\\ \hline
\end{tabular}\vspace{-1.0em}
\caption{Ablation study of TD, 3D Conv, TD+3D Conv, and TDC.}
\label{tab:tdc_ablation}
\end{table}

\subsubsection{Impact of TDC.}
As presented in Table~\ref{tab:tdc_ablation}, temporal difference (TD) alone suffers from limited recall (R) and AP\textsubscript{50}, as it leverages only frame-wise intensity variations while discarding most spatial contextual information. In contrast, 3D convolution produces lower F\textsubscript{1} and AP\textsubscript{50} due to its limited capability to model explicit temporal dependencies. Simply combining TD and 3D convolution yields some performance gains due to its reliance on single-scale spatio-temporal context. In contrast, our method achieves greater performance improvements, increasing AP\textsubscript{50} from 89.81 to 92.50, without introducing additional computational cost. This is because our proposed TDC fuses temporal difference and 3D convolution into a unified and learnable representation that captures multi-scale spatio-temporal context dependencies across different temporal ranges.

% As presented in Table~\ref{tab:tdc_ablation}, temporal difference (TD) alone suffers from limited R and $\text{AP}_{50}$, due to its weak spatial feature representation capacity. On the other hand, 3D convolution yields lower $\text{F}_1$ and $\text{AP}_{50}$ as it lacks explicit motion modeling. Combining the two still fails to fully resolve their individual limitations. In contrast, our proposed TDC achieves the highest P, R, $\text{F}_1$, and $\text{AP}_{50}$. These results demonstrate that the unified and learnable spatio-temporal representation in TDC not only captures diverse and comprehensive motion-contextual dependencies but also enables dynamic refinement of spatio-temporal features during training, without introducing additional computational cost.

\subsubsection{Impact of Different Spatio-Temporal Contextual Features.}
Table~\ref{tab:tdc_3_branch} shows that incorporating spatio-temporal contextual features at different temporal scales leads to notable performance gains. The S-TDC block improves P to 96.19 and F\textsubscript{1} to 94.91 by capturing fine-grained short-term spatio-temporal contextual features. M-TDC improves R to 95.79 and F\textsubscript{1} to 95.65. L-TDC captures long-range dependencies, achieving a P of 97.49 and an AP\textsubscript{50} of 92.35. When all branches are combined, the model reaches the best results across all metrics, confirming that multi-scale temporal modeling provides complementary motion cues essential for robust small target detection.
\begin{table}[!t]
\centering
\fontsize{9}{9.5}\selectfont
\begin{tabular}{ccccc}
\hline
Method & P & R & $\text{F}_1$ & $\text{AP}_{50}$ \\ \hline
w/o TDC & 86.70 & 88.04 & 87.36 & 75.97 \\
w/ S-TDC & 96.19 & 93.67 & 94.91 & 90.31 \\
w/ M-TDC & 95.51 & \underline{95.79} & 95.65 & 90.46 \\
w/ L-TDC & \underline{97.49} & 95.11 & \underline{96.29} & \underline{92.35} \\
w/ All & \textbf{97.61} & \textbf{95.93} & \textbf{96.76} & \textbf{92.50} \\ \hline
\end{tabular}\vspace{-0.6em}
\caption{Ablation study of different TDC blocks.}
\label{tab:tdc_3_branch}
\end{table}

\subsubsection{Impact of Re-parameterization in TDCR.}
According to Table~\ref{tab:re-parameterization}, re-parameterization reduces Params from 24.85M to 24.76M and FLOPs from 102.96G to 95.67G while maintaining consistent detection performance, demonstrating improved efficiency without sacrificing accuracy.
\begin{table}[!t]
\centering
\fontsize{9}{9.5}\selectfont
\begin{tabular}{ccccc}
\hline
Re-param & $\text{F}_1$ & $\text{AP}_{50}$ & Params (M) & FLOPs (G)\\ \hline
w/o & 96.76 & 92.50 & 24.85 & 102.96 \\ 
w/ & 96.76 & 92.50 & \textbf{24.76} & \textbf{95.67} \\ \hline
\end{tabular}\vspace{-0.6em}
\caption{Ablation study of re-parameterization in TDCR.}
\label{tab:re-parameterization}
\end{table}

\subsubsection{Impact of TDCSTA.}
From Table~\ref{tab:sta_ablation}, we observe that setting temporal difference convolution features (TDCF) as query, with spatio-temporal features (STF) and spatial features (SF) as key and value, yields the best performance with F\textsubscript{1} of 96.74 and AP\textsubscript{50} of 92.35. Replacing the query with STF or SF leads to noticeable drops in all metrics, confirming that TDCF provides more discriminative guidance in TDCSTA, which is essential for accurate target localization in cluttered scenes.
\begin{table}[!t]
\centering
\fontsize{9}{10}\selectfont %font size and line height
\begin{tabular}{ccccccc}
\hline
Query & Key & Value & P & R & $\text{F}_1$ & $\text{AP}_{50}$ \\ \hline
TDCF & STF & SF & \textbf{97.53} & \textbf{95.96} & \textbf{96.74} & \textbf{92.35} \\
STF & TDCF & SF & \underline{93.11} & \underline{90.15} & \underline{91.61} & \underline{87.22} \\
SF & STF & TDCF & 92.84 & 89.73 & 91.26 & 86.39 \\ \hline
\end{tabular}\vspace{-1.0em}
\caption{Ablation study of the TDCSTA module with different combinations of query, key, and value in the cross-attention mechanism.}
\label{tab:sta_ablation}
\end{table}

\section{Conclusion}
This paper introduces a novel model TDCNet for moving IRSTD. TDCNet incorporates two key designs: the TDCR module and the TDCSTA mechanism. TDCR module captures multi-scale temporal contextual features while suppressing complex backgrounds without incurring additional computational cost during inference. TDCSTA mechanism models semantic relationships between two 3D feature streams to refine the representation of critical target regions in the current frame. These components effectively enhance spatio-temporal feature representation, enabling TDCNet to outperform existing methods on the IRSTD-UAV and public IRDST datasets. Despite its strong performance, TDCNet still has relatively high model complexity, which we plan to address in future work by exploring lightweight and efficient architectures.

%We proposed FACT that performs temporal modeling on frame and action levels in parallel and performs iterative bidirectional information transfer between them for frame and action feature refinement. FACT has a frame branch to learn frame features with convolution, an action branch to learn action tokens with transformer, and cross-attentions for cross-branch communication. By extensive experiments, we showed FACT exceeds all prior methods on four datasets with a low inference complexity, and is able to also incorporate textual transcripts when they are available.

%\newpage
\section{Acknowledgments}
This work was supported by the Open Research Fund of the National Key Laboratory of Multispectral Information Intelligent Processing Technology under Grant 61421132301 and was supported in part by the projects of the National Natural Science Foundation of China under Grants No. 62371203 and 62301228.

%\bigskip

\bibliography{aaai2026}

% 主文档最后
\clearpage
\def\INSIDEMAIN{} % 标记：当前是被主文档包含

% Title

% Your title must be in mixed case, not sentence case.
% That means all verbs (including short verbs like be, is, using,and go),
% nouns, adverbs, adjectives should be capitalized, including both words in hyphenated terms, while
% articles, conjunctions, and prepositions are lower case unless they
% directly follow a colon or long dash
%\title{Spatio-Temporal Context Learning with Temporal Difference Convolution for Moving Infrared Small Target Detection\\ \noindent{\bf{\fontsize{20}{25}\selectfont\textbf{------}}}Supplementary Material\noindent\bf{\fontsize{20}{25}\selectfont\textbf{------}}}

\providecommand{\suppTitle}{Supplementary Material}

\ifdefined\INSIDEMAIN
  % 被主文件包含时走这里
  \twocolumn[
  \centering\fontsize{25}{10}\selectfont\section*{\suppTitle} % \fontsize{14}{16}%size and linespace
   \vspace{1em}% 调整这个值来控制空行距离
]
  
\else
  % 单独编译时走这里
  \title{\suppTitle}
  \maketitle
\fi

%%%% 确保补充材料重新编号
\setcounter{section}{0}  % 重置章节编号
\setcounter{subsection}{0}  % 重置小节编号
\setcounter{subsubsection}{0}  % 重置子小节编号

%%%% 确保补充材料中图和表重新编号
\setcounter{figure}{0}   % 重置图的编号
\setcounter{table}{0}    % 重置表格的编号
%\renewcommand{\thefigure}{\arabic{figure}} % 图的编号前缀添加“S”
%\renewcommand{\thetable}{\arabic{table}}   % 表格的编号前缀添加“S”

% ---- 补充内容从这里开始 ----

\section{Overview}
% In this part, we provide the supplementary material of our work that are not included in the article due to the space limitations. 
In this part, we provide additional details and more experimental results to further validate the effectiveness of the proposed temporal difference convolution (TDC) network (TDCNet). The structure of this supplementary material is organized as follows.
% \begin{itemize}
% \item  We first introduce the details of two real infrared Unmanned Aerial Vehicle (UAV)  and Car datasets that we use to train and test our model ECFNet.
% \item  We then compare our method with state-of-the-art (SOTA) methods on two real infrared datasets.
% \item We also analysis more details about our method like the threshold $\tau$ in the metrics CRR and CRP, the difference between our DAT and the denoising training method in the DN-DETR. 
% \item  We further conduct more ablation studies to validate the impact of the main module and the key hyper-parameters inside the proposed methods. 
% \end{itemize}  
\begin{itemize}
\item \textbf{Further details of the proposed short-term TDC (S-TDC) and mid-term TDC (M-TDC) blocks.} Due to space constraints in the main paper, we only presented details of the long-term TDC (L-TDC) block. In Section~\ref{sec:tdcnet}, we provide additional construction details of the proposed S-TDC and M-TDC blocks, along with further architectural analysis. We also elaborate on the implementation details, including the progressive training strategy.

\item \textbf{Further details of our self-constructed IRSTD-UAV and the public IRDST datasets.} In Section~\ref{sec:datasets}, we provide a detailed description of our self-constructed IRSTD-UAV and the public IRDST~\cite{2023IRDST} datasets, including their target scale distribution, background diversity, and dataset division strategy.
\item \textbf{Further qualitative results.} In Section~\ref{sec:qualitative}, we present additional qualitative comparisons of TDCNet with a wide range of state-of-the-art (SOTA) methods on both the IRSTD-UAV and IRDST datasets, along with precision–recall curve analyses for quantitative evaluation.
\item \textbf{Further ablation studies.} In Section~\ref{sec:ablation}, we investigate the impact of the temporal kernel size in the TDC block and the number of input frames on detection performance and computational efficiency, demonstrating optimal trade-offs between accuracy and resource consumption.
\end{itemize}

%This supplementary material highlights the robustness, efficiency, and effectiveness of the TDCNet across a wide range of experimental scenarios, providing strong support for the claims made in the main paper.

\begin{figure}[!t]
\centering
\includegraphics[width=3.3in,keepaspectratio]{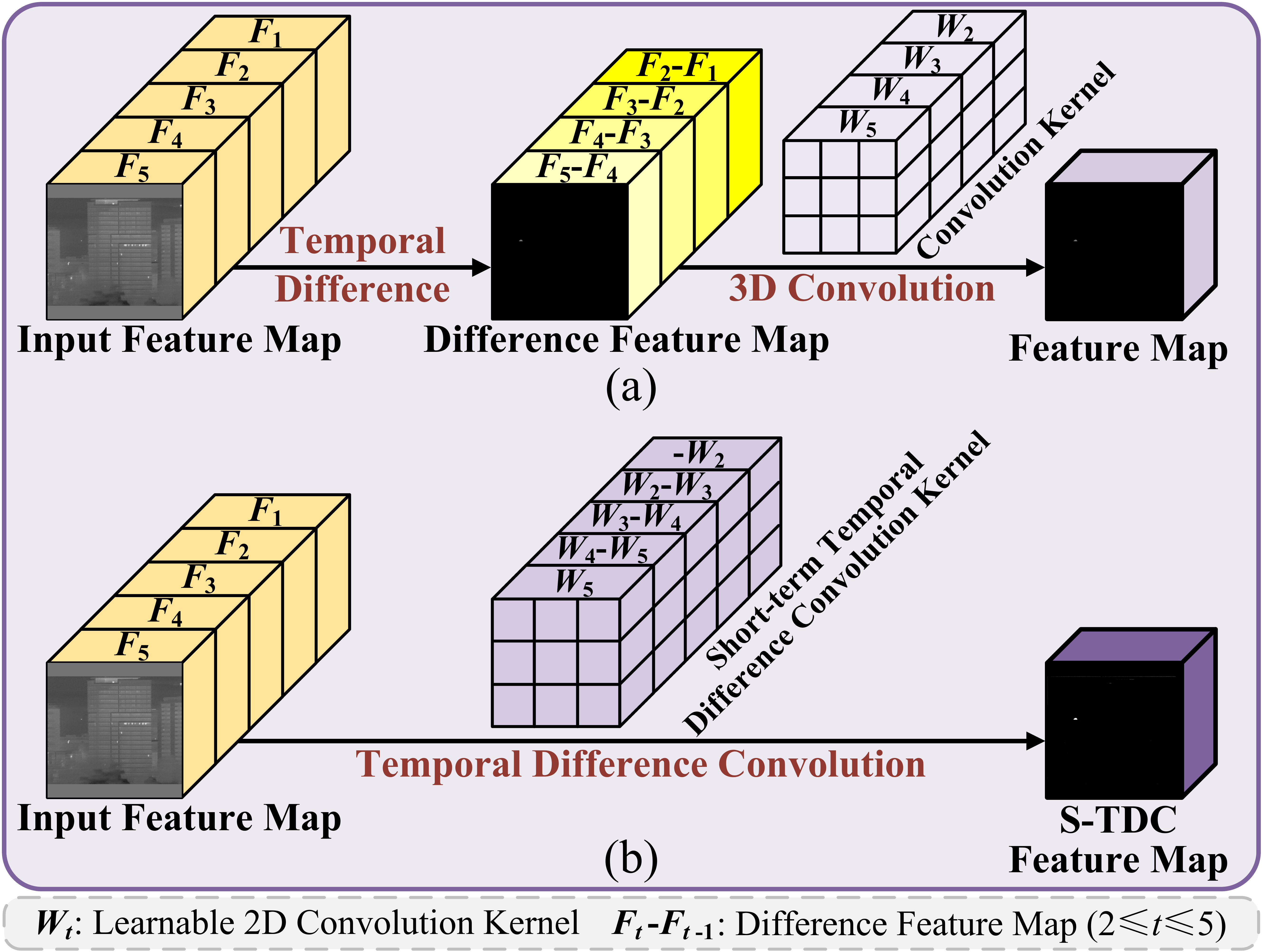} 
\caption{(a) Sequential combination of existing temporal difference and 3D convolution, and (b) Our proposed short-term temporal difference convolution (S-TDC) block, which fuses temporal difference modeling and 3D convolution into a unified spatio-temporal convolution representation.}
\label{fig:stdc}
\end{figure}

\begin{figure}[!t]
\centering
\includegraphics[width=3.3in,keepaspectratio]{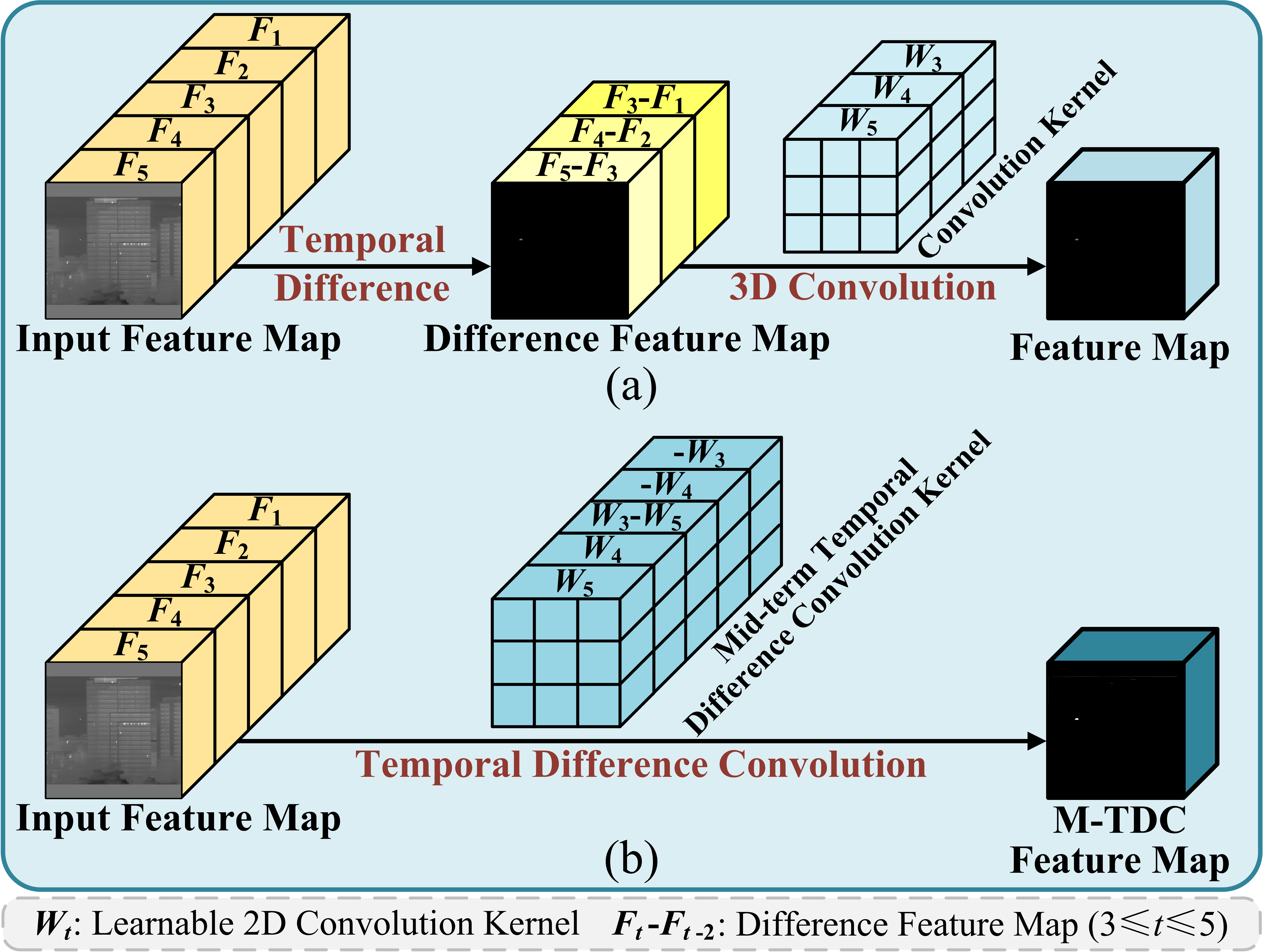} 
\caption{(a) Sequential combination of existing temporal difference and 3D convolution, and (b) Our proposed mid-term temporal difference convolution (M-TDC) block, which fuses temporal difference modeling and 3D convolution into a unified spatio-temporal convolution representation.}
\label{fig:mtdc}
\end{figure}

\section{Further Details and Analysis of Our TDCNet}
\label{sec:tdcnet}
\subsection{Further Details of the Proposed S-TDC and M-TDC Blocks}
Due to space constraints in the main paper, we only present details of the long-term TDC (L-TDC) block. In this section, we provide further details of the proposed short-term TDC (S-TDC) and mid-term TDC (M-TDC) blocks here. The S-TDC block focuses on capturing fine-grained and fast-changing motion patterns by computing temporal differences between consecutive frames. This design enhances the network’s sensitivity to subtle temporal variations, enabling more precise modeling of short-term motion dynamics. Figure~\ref{fig:stdc} (b) illustrates our S-TDC block, which fuses temporal difference modeling and 3D convolution into a unified spatio-temporal convolution representation, differing from the sequential combination of existing temporal difference and 3D convolution shown in Figure~\ref{fig:stdc} (a). Similarly, the mid-term TDC (M-TDC) block is designed to capture intermediate temporal dependencies by computing differences between frames spaced two intervals apart, as illustrated in Figure~\ref{fig:mtdc} (b). This intermediate temporal modeling complements the S-TDC and L-TDC blocks by effectively capturing motion contextual dependencies at a mid-range temporal scale while mitigating noise and redundant motion information. The M-TDC block further strengthens the overall spatio-temporal feature representation of our method.

\subsection{Training Process of the TDCNet}
We adopt a progressive training strategy to ensure effective spatio-temporal feature learning in our TDCNet. First, the 2D backbone is pre-trained on single-frame infrared images to extract robust spatial features. In parallel, the 3D backbone is pre-trained on video clips to learn general spatio-temporal patterns across multiple frames. Once both backbones are sufficiently trained, their parameters are frozen to prevent overfitting and to stabilize the subsequent optimization. In the final stage, we train the proposed TDC backbone together with the TDC-guided spatio-temporal attention (TDCSTA) module on infrared video sequences. This staged approach allows each component to focus on learning its respective representation, ultimately enhancing the network’s capability to model multi-scale spatio-temporal context in complex infrared scenarios.

\begin{figure*}[!t]
\centering
\includegraphics[width=7in,keepaspectratio]{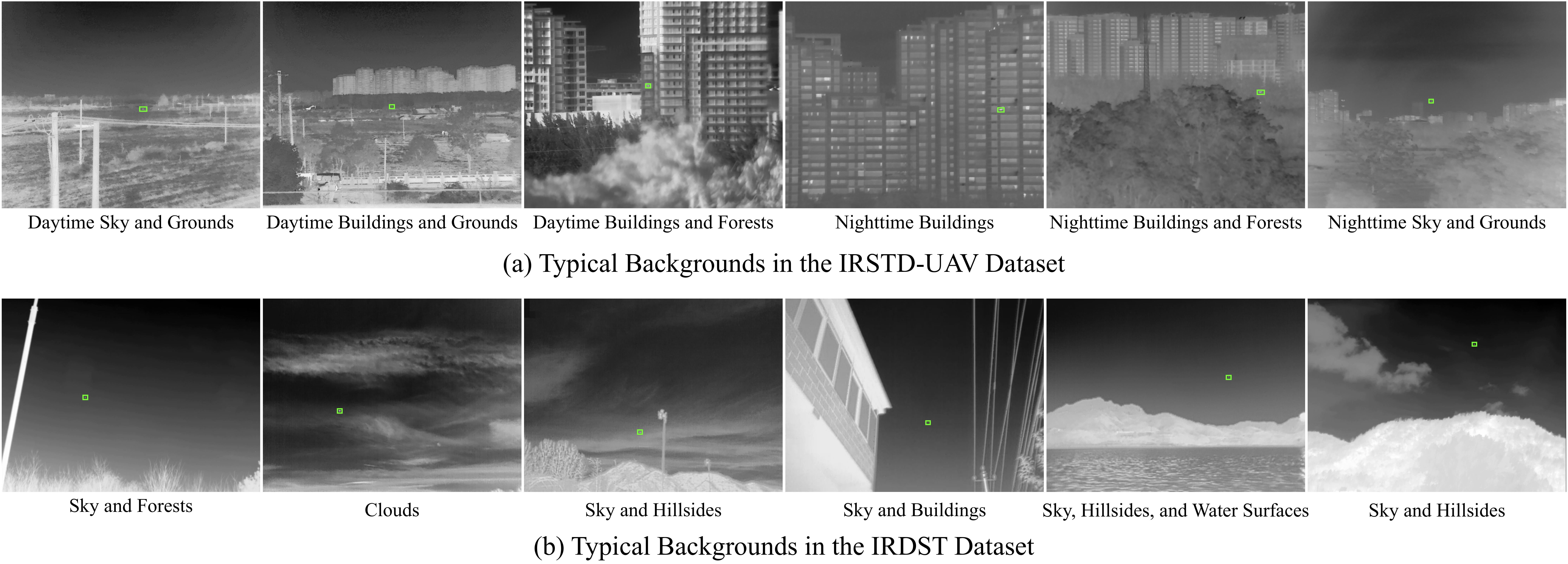} 
\caption{Typical backgrounds in the IRSTD-UAV and IRDST datasets.}
\label{fig:vis_supp_background}
\end{figure*}

\begin{figure}[!t]
\centering
\includegraphics[width=3.3in,keepaspectratio]{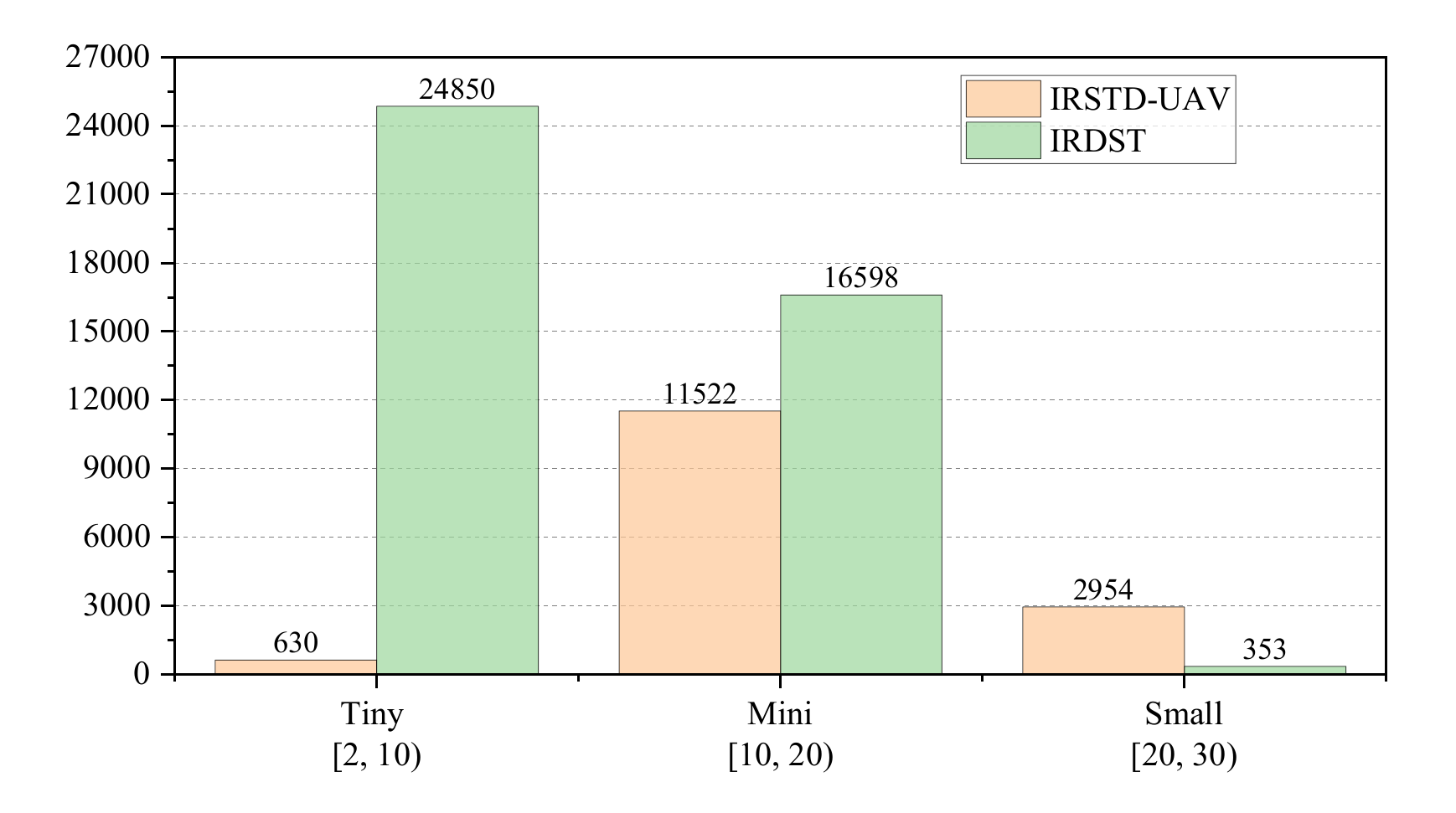} 
\caption{Statistics of target scales in the IRSTD-UAV and IRDST datasets.}
\label{fig:tgt_scale}
\end{figure}

\section{Further Details of the IRSTD-UAV and IRDST Datasets}
\begin{table*}[!t]
\centering
\fontsize{7.2}{11}\selectfont %font size and line height
\setlength\tabcolsep{3pt}
\begin{tabular}{ccccccccc}
\hline
Dataset & Annotations & Sequential & Classes & Target Size & Target Type & Sequences & Frame Number & Image Size \\
\hline
IRSTD-UAV & Bounding boxes & \checkmark & 1 & 24 $\sim$ 437 pixels & Unmanned Aerial Vehicles & 17 & 15,106 & 640 $\times$ 512 \\
IRDST     & Bounding boxes & \checkmark & 1 & 25 $\sim$ 396 pixels & Helicopter, Airplane, Drone & 92 & 40,656 & 720 $\times$ 480, 934 $\times$ 696, 992 $\times$ 742 \\
\hline
\end{tabular}
\caption{Details of the IRSTD-UAV and IRDST datasets.}
\label{table:datasets}
\end{table*}
\label{sec:datasets}
\subsection{Statistics and Analysis of the IRSTD-UAV and IRDST Datasets}

\subsubsection{Scale distribution of targets in two datasets.}
As shown in Figure~\ref{fig:tgt_scale}, in the IRSTD-UAV dataset, tiny-scale targets (Tiny, [2, 10)) account for approximately 4.17\% (630 targets), mini-scale targets (Mini, [10, 20)) account for about 76.27\% (11,522 targets), and small-scale targets (Small, [20, 30)) represent roughly 19.56\% (2,954 targets). In conclusion, the IRSTD-UAV dataset is entirely composed of small targets.

The IRDST dataset comprises approximately 59.45\% tiny-scale targets (24,850 targets), 39.71\% mini-scale targets (16,598 targets), and 0.84\% small-scale targets (353 targets).

\subsubsection{Diversity of backgrounds.} Figure~\ref{fig:vis_supp_background} (a) shows representative background images from the IRSTD-UAV dataset. These scenes cover two distinct lighting conditions (daytime and nighttime) and feature a diverse range of backgrounds, including sky, grounds, buildings, and forests. The selected real-world scenes effectively capture complex environmental variations, providing a comprehensive and challenging testbed for evaluating the robustness of our proposed TDCNet against cluttered infrared backgrounds.

Figure~\ref{fig:vis_supp_background} (b) presents typical background images from the IRDST dataset, which include scenes with sky, forests, clouds, hillsides, buildings, and water surfaces. Compared to IRDST, the backgrounds in IRSTD-UAV exhibit greater complexity and diversity, posing more significant challenges for small target detection.

\subsection{Dataset Division Strategy}
The IRSTD-UAV dataset consists of 17 sequences with a total of 15,106 frames. Each sequence is first divided into segments of 50 consecutive frames. Subsequently, the segments are randomly shuffled and split into training and validation sets at an 8:2 ratio.

The IRDST dataset contains 92 video sequences totaling 40,656 frames. Following the official split, 42 videos comprising 20,398 frames are used for training, and 43 videos comprising 20,258 frames for validation.

%The IRSTD-UAV dataset is available for download at IRSTD-\citet{UAV}.

\begin{figure*}[!t]
\centering
\includegraphics[width=7in,keepaspectratio]{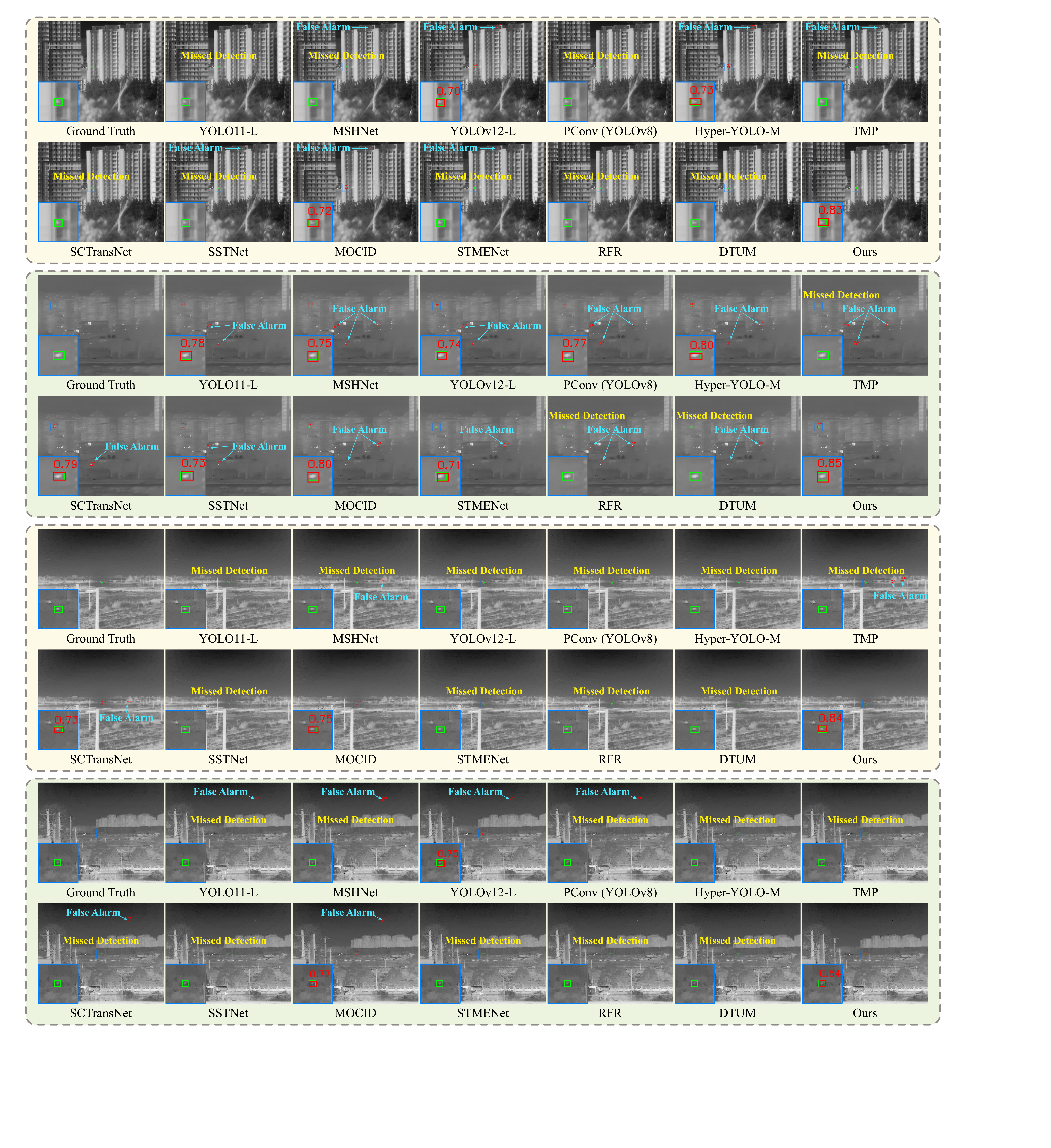} 
\caption{Qualitative comparison of our TDCNet and SOTA methods on the IRSTD-UAV dataset. Zoomed-in views are shown in the bottom-left corner. Boxes in green and red represent ground-truth and detected targets, respectively.}
\label{fig:vis_supp_uav}
\end{figure*}

\begin{figure*}[!t]
\centering
\includegraphics[width=7in,keepaspectratio]{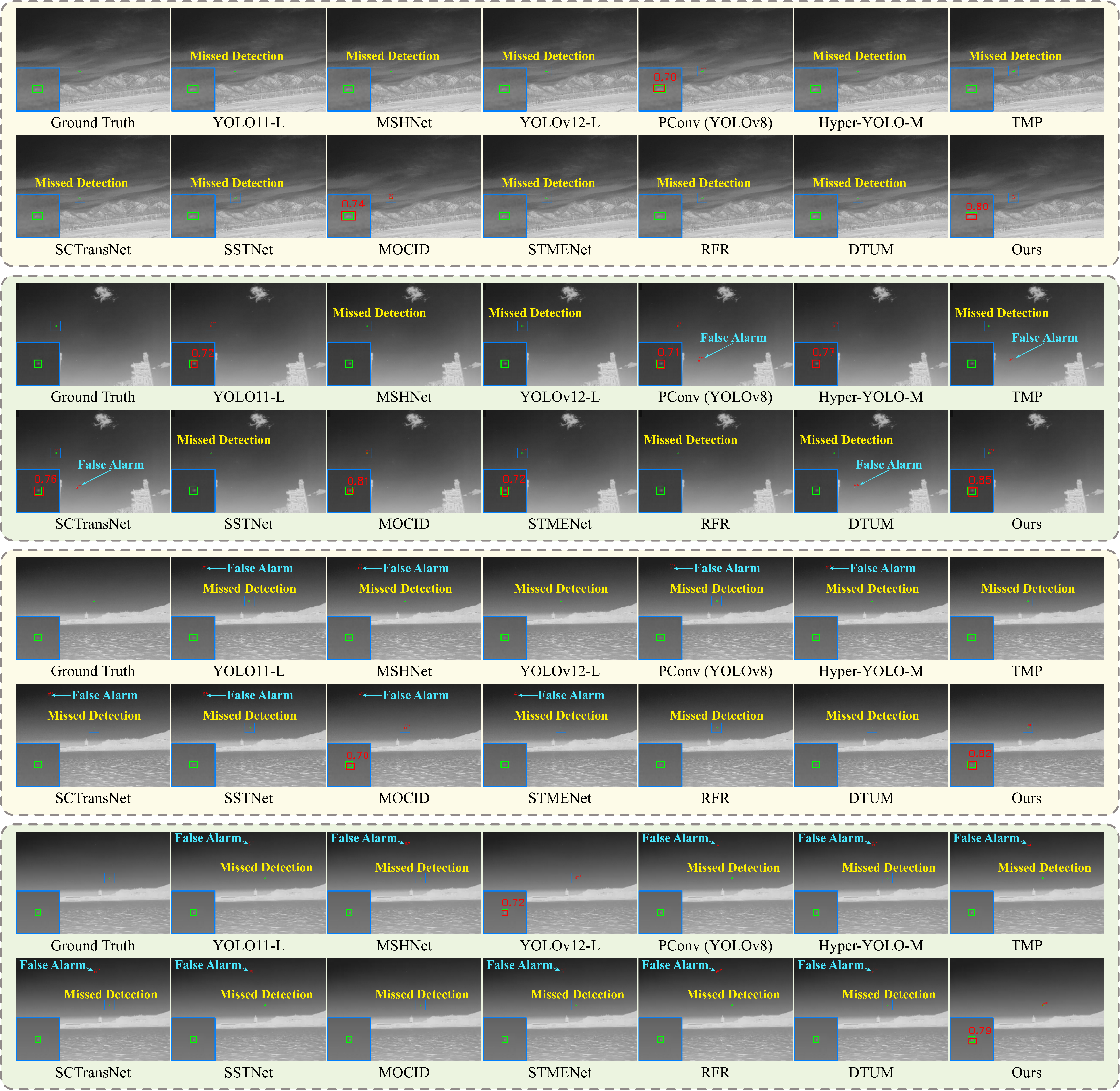} 
\caption{Qualitative comparison of our TDCNet and SOTA methods on the IRDST dataset. Zoomed-in views are shown in the bottom-left corner. Boxes in green and red represent ground-truth and detected targets, respectively.}
\label{fig:vis_supp_irdst}
\end{figure*}

\section{Further Qualitative Comparisons of TDCNet with SOTA Methods}
\label{sec:qualitative}
To comprehensively evaluate the robustness and generalization of our method in challenging infrared scenarios, we further compare TDCNet with a wide range of SOTA methods on the IRSTD-UAV and IRDST datasets. These methods include three general-purpose convolutional neural network (CNN)-based single-frame methods (YOLO11-L~\cite{Jocher_Ultralytics_YOLO11}, YOLOv12-L~\cite{tian2025yolov12}, and Hyper-YOLO-M~\cite{feng2024hyperyolo}), two infrared-specific CNN-based single-frame methods (MSHNet~\cite{2024CVPRLiuMSHNet} and PConv (YOLOv8)~\cite{2025AAAIYangPinwheelConv}), one infrared-specific transformer-based method (SCTransNet~\cite{2024TGRSYuanSCTransNet}), and six infrared-specific CNN-based multi-frame methods (TMP~\cite{2024ZhuESATMP}, SSTNet~\cite{2024TGRSChenSSTNet}, MOCID~\cite{2025AAAIZhangMOCID}, STMENet~\cite{2025EAAIPengSTME}, RFR~\cite{Ying_2025RFR}, and DTUM~\cite{2025TNNLSLiDTUM}). We first present additional qualitative results of these methods on both datasets, followed by precision–recall (P–R) curves on the IRSTD-UAV dataset for quantitative comparison.

\subsection{Further Qualitative Results on the IRSTD-UAV Dataset}
As shown in Figure~\ref{fig:vis_supp_uav}, we provide additional qualitative results on the IRSTD-UAV dataset, which features a diverse range of challenging scenarios. These include brightly lit buildings under strong daylight conditions, nighttime scenes where numerous lights introduce significant visual distractions, and urban environments cluttered with structures and textures. Such complex backgrounds pose considerable challenges for IRSTD. Despite incorporating infrared-specific designs or temporal modeling strategies, most compared methods struggle to deliver consistent performance under these conditions, especially in scenes with dense structures, severe background noise, or subtle target-background contrast. Specifically:
\begin{itemize}
    \item YOLO11-L, YOLOv12-L, and Hyper-YOLO-M, as general-purpose CNN-based methods, frequently suffer from false alarms in cluttered environments. Without explicit temporal modeling or infrared-specific feature design, these methods often misidentify bright background textures (e.g., building edges or artificial lights) as targets, especially under complex nighttime conditions.

    \item MSHNet and PConv (YOLOv8), though specifically designed for IRSTD, still struggle under low-contrast or noisy scenarios. Their strong reliance on appearance-based cues makes them vulnerable to target-background ambiguity, leading to missed detections particularly in urban and ground scenes.

    \item TMP, SSTNet, and STMENet, which utilize temporal modeling, can capture motion patterns but are often misled by fluctuating background elements such as light flicker or reflections on surfaces. This results in frequent false alarms in dynamic scenes with strong background interference.
    
    \item SCTransNet, while benefiting from transformer-based global feature modeling, lacks explicit motion-awareness. As a result, it often produces false alarms in complex urban backgrounds where the target features are easily confused with complex background interference.

    \item MOCID, RFR, and DTUM attempt to model motion context but lack mechanisms to handle multi-scale temporal dependencies or suppress background clutter effectively. Consequently, they frequently fail in scenes with dense distractors or weak target signals.
\end{itemize}

In contrast, our TDCNet demonstrates superior robustness in such challenging scenarios. Even under strong background clutter, such as urban structures and light-like distractors, TDCNet effectively highlights true UAV targets while suppressing false alarms. This improvement stems from the temporal difference convolution re-parameterization (TDCR) module's capability to capture multi-scale motion-contextual dependencies, while the TDCSTA selectively enhances target-relevant features and suppresses irrelevant background interference.

\subsection{Further Qualitative Results on the Public IRDST Dataset}
As presented in Figure~\ref{fig:vis_supp_irdst}, we provide additional qualitative comparisons on the IRDST dataset, which primarily features scenarios with relatively clean backgrounds such as open sky, clouds, hillsides, and water surfaces. While these scenes exhibit fewer structural distractions, infrared targets tend to be extremely small, dim, or embedded within low-emissivity regions, posing distinct challenges for detection. Many existing methods fail to maintain reliable performance under these conditions due to limited sensitivity to subtle spatio-temporal variations. Specifically:
\begin{itemize}
    \item YOLO11-L, YOLOv12-L, and Hyper-YOLO-M struggle to detect faint or low-contrast targets in IRDST scenes. Their CNN-based design lacks fine-grained temporal sensitivity and infrared-specific enhancements, leading to frequent missed detections when targets appear against smooth or gradually varying backgrounds.

    \item MSHNet and PConv (YOLOv8), despite being tailored for IRSTD, still rely heavily on appearance-based contrast. This makes them prone to overlooking dim targets that blend into the sky or cloud, especially when the thermal gradient is minimal.

    \item TMP, SSTNet, and STMENet benefit from temporal modeling, but their relatively shallow motion cues are insufficient for distinguishing slow-moving or stationary small objects. They often fail to detect targets under weak spatio-temporal contrast or in the presence of environmental noise.

    \item SCTransNet, while capable of capturing global context, lacks the precision to isolate weak signals from broad, smooth thermal backgrounds. It frequently fails to highlight tiny or low-salience targets due to inadequate motion modeling capabilities.

    \item MOCID, RFR, and DTUM incorporate motion modeling but show limited effectiveness when motion cues are subtle and background interference, though minimal, is still sufficient to obscure weak target signatures. Their insufficient multi-scale temporal discrimination often results in unstable detections.
\end{itemize}

In contrast, our TDCNet effectively integrates multi-scale motion-contextual modeling through the TDCR module with selective spatio-temporal feature enhancement enabled by the TDCSTA mechanism. This powerful combination enables robust detection of extremely weak targets that are often embedded within relatively simple yet thermally ambiguous backgrounds such as sky, clouds, mountains, and water surfaces. By accurately capturing subtle motion cues and suppressing irrelevant background fluctuations, our approach consistently outperforms SOTA methods that struggle to distinguish faint targets from smooth or low-contrast infrared scenes, demonstrating superior generalization and robustness across diverse infrared detection challenges.

\begin{figure}[!t]
\centering
\includegraphics[width=3.3in,keepaspectratio]{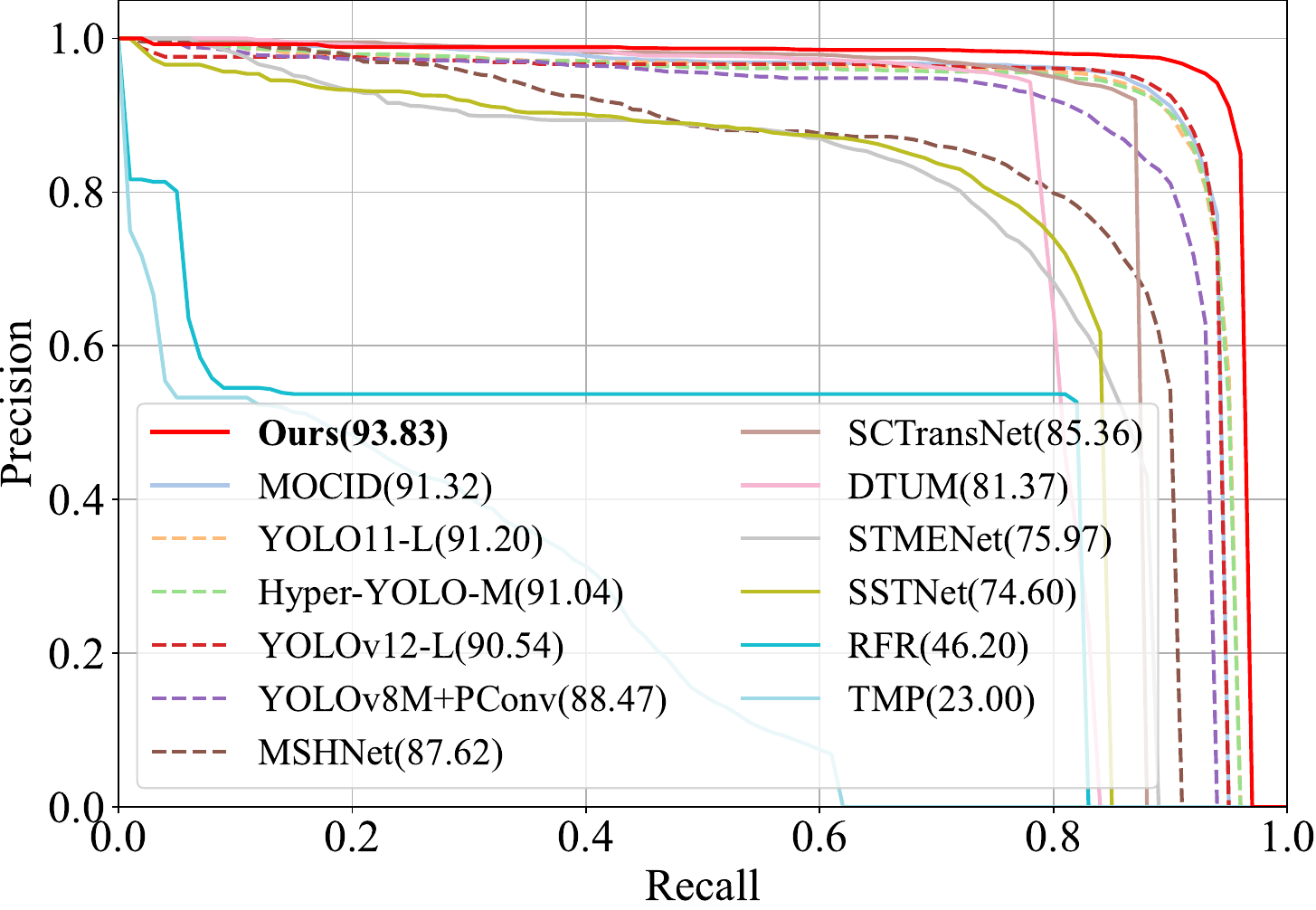} 
\caption{Precision–recall (P–R) curves of our method and other IRSTD methods on the IRSTD-UAV dataset. The area under each curve is indicated after the corresponding method name. Dashed lines represent single-frame methods, while solid lines denote multi-frame methods.}
\label{fig:pr_curves}
\end{figure}

\subsection{Precision–Recall Curve Comparison on the IRSTD-UAV Dataset}
We plot the precision–recall curves of our TDCNet alongside those of other detection methods on the IRSTD-UAV dataset, as shown in Figure~\ref{fig:pr_curves}. Different colors are used to clearly distinguish each method for better visual comparison. A larger area under the curve (AUC) reflects better detection performance. As illustrated in the figure, our TDCNet achieves the highest AUC among all SOTA methods, demonstrating its superior detection capability.

\section{Further Ablation Studies}
\label{sec:ablation}
\subsection{Impact of Temporal Kernel Size in TDC}
\begin{table}[!t]
\centering
\fontsize{9}{10}\selectfont %font size and line height
\begin{tabular}{ccccc}
\hline
\makecell[c]{Kernel Size \\ ($K \times 3 \times 3$)} & $\text{F}_1$ & $\text{AP}_{50}$ & Params (M) & FLOPs (G) \\ \hline
$K = 3$ & 95.13 & 91.24 & \textbf{10.335} & \textbf{44.762} \\
$K = 5$ & \textbf{96.76} & \textbf{92.50} & \underline{10.385} & \underline{45.749} \\
$K = 7$ & \underline{95.64} & \underline{91.71} & 10.413 & 46.735 \\ \hline
\end{tabular}
\caption{Ablation study of temporal kernel size in TDC.}
\label{tab:tdc_kernel}
\end{table}
Table~\ref{tab:tdc_kernel} shows that increasing the temporal kernel size \(K\) from 3 to 5 leads to improved performance, with \(\text{F}_1\) increasing from 95.13 to 96.76 and \(\text{AP}_{50}\) from 91.24 to 92.50. A kernel size of 5 provides the best balance between accuracy and computational efficiency. Further increasing \(K\) to 7 does not yield significant performance gains but results in higher computational cost.

\subsection{Impact of Input Frame Number}
\begin{table}[!t]
\centering
\fontsize{9}{10}\selectfont %font size and line height
\setlength\tabcolsep{2.5pt} % 调整列间距
\begin{tabular}{cccccccc}  
\hline
$T$ & 3 & 4 & 5 & 6 & 7 & 8 & 9 \\ \hline
$\text{F}_1$ & 96.74 & 96.10 & \textbf{97.12} & 96.19 & \underline{96.91} & 96.23 & 96.42 \\
$\text{AP}_{50}$ & 92.48 & 91.68 & \textbf{93.83} & 91.86 & \underline{93.24} & 91.72 & 91.97 \\
FLOPs (G) & \textbf{81.3} & \underline{82.7} & 95.7 & 97.0 & 98.5 & 99.9 & 132.1 \\
FPS & \textbf{19.8} & \underline{19.4} & 18.5 & 18.2 & 17.4 & 16.9 & 16.4 \\ \hline
\end{tabular}
\caption{Ablation study of input frame number ($T$).}
\label{tab:input_frame_ablation}
\end{table}
Table~\ref{tab:input_frame_ablation} illustrates that using 5 frames achieves the best trade-off between accuracy, with $\text{F}_1$ reaching 97.12 and $\text{AP}_{50}$ reaching 93.83, and efficiency. While using more frames yields comparable accuracy, it increases computational cost; using fewer frames reduces cost but leads to lower performance.

\subsection{Robustness to Static and Slowly Moving Targets}
\begin{table}[!t]
\centering
% \fontsize{9}{10}\selectfont
% \setlength\tabcolsep{3pt}
\begin{tabular}{ccccc}
\hline
P & R & F$_1$ & AP$_{50}$ \\ \hline
99.32 & 97.20 & 98.25 & 96.27 \\ \hline
\end{tabular}
\caption{Performance of TDCNet on static or extremely slow-moving infrared targets.}
\label{tab:static_target}
\end{table}

Table~\ref{tab:static_target} shows that our model maintains high accuracy even for static or extremely slow-moving targets (5 sequences, 50 frames each, total 250 frames), demonstrating strong robustness beyond the primary moving target detection task.

\subsection{Robustness under Inter-frame Background Misalignment}
\begin{table}[!t]
\centering
\fontsize{9}{10}\selectfont
\begin{tabular}{ccccc}
\hline
Inter-frame Interval & P & R & F$_1$ & AP$_{50}$ \\ \hline
0 & 97.99 & 96.27 & 97.12 & 93.83 \\
5 & 97.61 & 93.85 & 95.69 & 91.28 \\
10 & 96.74 & 92.43 & 94.54 & 89.62 \\
15 & 95.10 & 91.87 & 93.46 & 86.73 \\
20 & 93.69 & 90.25 & 91.93 & 83.12 \\ \hline
\end{tabular}
\caption{Performance under inter-frame background misalignment caused by high-speed UAV motion. The inter-frame interval indicates the number of frames skipped to simulate increasing misalignment.}
\label{tab:frame_misalignment}
\end{table}

Table~\ref{tab:frame_misalignment} indicates that even under mild to moderate inter-frame background misalignment, our method maintains strong performance. Only in extreme misalignment scenarios does the accuracy noticeably degrade, showing the method's robustness to most real-world UAV motion conditions.
%\appendix

%\bigskip

\bibliography{aaai2026}
%\end{document}

\end{document}